\documentclass[journal]{IEEEtran}
\usepackage{ifpdf}

\usepackage{cite}

\ifCLASSINFOpdf
  \usepackage[pdftex]{graphicx}
  \graphicspath{{../pdf/}{../jpeg/}}
  \DeclareGraphicsExtensions{.pdf,.jpeg,.png}
\else
  \usepackage[dvips]{graphicx}
  \graphicspath{{../eps/}}
  \DeclareGraphicsExtensions{.eps}
\fi
\usepackage{amsmath}
\usepackage{amsthm, amssymb}
\usepackage{mathrsfs}
\usepackage{booktabs}
\usepackage{multirow}
\usepackage{multicol}
\usepackage{makecell}

\usepackage[utf8]{inputenc}

\usepackage{algorithmic}
\usepackage{algorithm}
\makeatletter
\newcommand{\removelatexerror}{\let\@latex@error\@gobble}
\makeatother

\usepackage{array}

\ifCLASSOPTIONcompsoc
  \usepackage[caption=false,font=normalsize,labelfont=sf,textfont=sf]{subfig}
\else
  \usepackage[caption=false,font=footnotesize]{subfig}
\fi

\ifCLASSOPTIONcaptionsoff
  \usepackage[nomarkers]{endfloat}
 \let\MYoriglatexcaption\caption
 \renewcommand{\caption}[2][\relax]{\MYoriglatexcaption[#2]{#2}}
\fi

\usepackage{verbatim}

\usepackage{url}

\begin{document}
%
\title{Discriminative Feature Representaion with Spatio-temporal Cues for Vehicle Re-identification}
%
%
%

\author{Jingzheng~Tu,
	Cailian~Chen,~\IEEEmembership{Member,~IEEE},
		Xiaolin~Huang,~\IEEEmembership{Member,~IEEE},
		Jianping~He,~\IEEEmembership{Senior Member,~IEEE},
		Xinping~Guan,~\IEEEmembership{Fellow,~IEEE}
\thanks{J. Tu, C. Chen, X. Huang, J. He and X. Guan are with the Department of Automation, Shanghai Jiao Tong University, Shanghai 200240, China and also with the Key Laboratory of System Control and Information Processing, Ministry of Education of China, Shanghai Jiao Tong University, Shanghai 200240, China (e-mail: tujingzheng@sjtu.edu.cn, cailianchen@sjtu.edu.cn, xiaolinhuang@sjtu.edu.cn, jphe@sjtu.edu.cn, xpguan@sjtu.edu.cn).
	} 
}

%
%

\markboth{Journal of \LaTeX\ Class Files,~Vol.~14, No.~8, August~2015}%
{Shell \MakeLowercase{\textit{et al.}}: Bare Demo of IEEEtran.cls for IEEE Journals}
%



\maketitle

\begin{abstract}
Vehicle re-identification (re-ID) aims to discover and match the target vehicles from a gallery image set taken by different cameras on a wide range of road networks. It is crucial for lots of applications such as security surveillance and traffic management. The remarkably similar appearances of distinct vehicles and the significant changes of viewpoints and illumination conditions take grand challenges to vehicle re-ID. Conventional solutions focus on designing global visual appearances without sufficient consideration of vehicles' spatio-tamporal relationships in different images. 
In this paper, we propose a novel discriminative feature representation with spatio-temporal clues (DFR-ST) for vehicle re-ID. It is capable of building robust features in the embedding space by involving appearance and spatio-temporal information. Based on this multi-modal information, the proposed DFR-ST constructs an appearance model for a multi-grained visual representation by a two-stream architecture and a spatio-temporal metric to provide complementary information. 
Experimental results on two public datasets demonstrate DFR-ST outperforms the state-of-the-art methods, which validate the effectiveness of the proposed method.

\end{abstract}


%
\IEEEpeerreviewmaketitle

\section{Introduction}

\IEEEPARstart{W}{ith} increasing demand for public security and the rapid growth of vehicles, vehicle re-identification (re-ID) has become one of the most pivotal technologies for intelligent urban surveillance. It also has a wide range of potential applications including multi-target multi-camera tracking, traffic flow modeling \cite{he2020multi,bashir2019vrproud, castaneda2016scalable}.
The main task of vehicle re-ID is to locate vehicles accurately and identify the same vehicle over multiple cameras with particular perspectives.  

\begin{figure}[!t]
	\centering
	\includegraphics[width=2.5in]{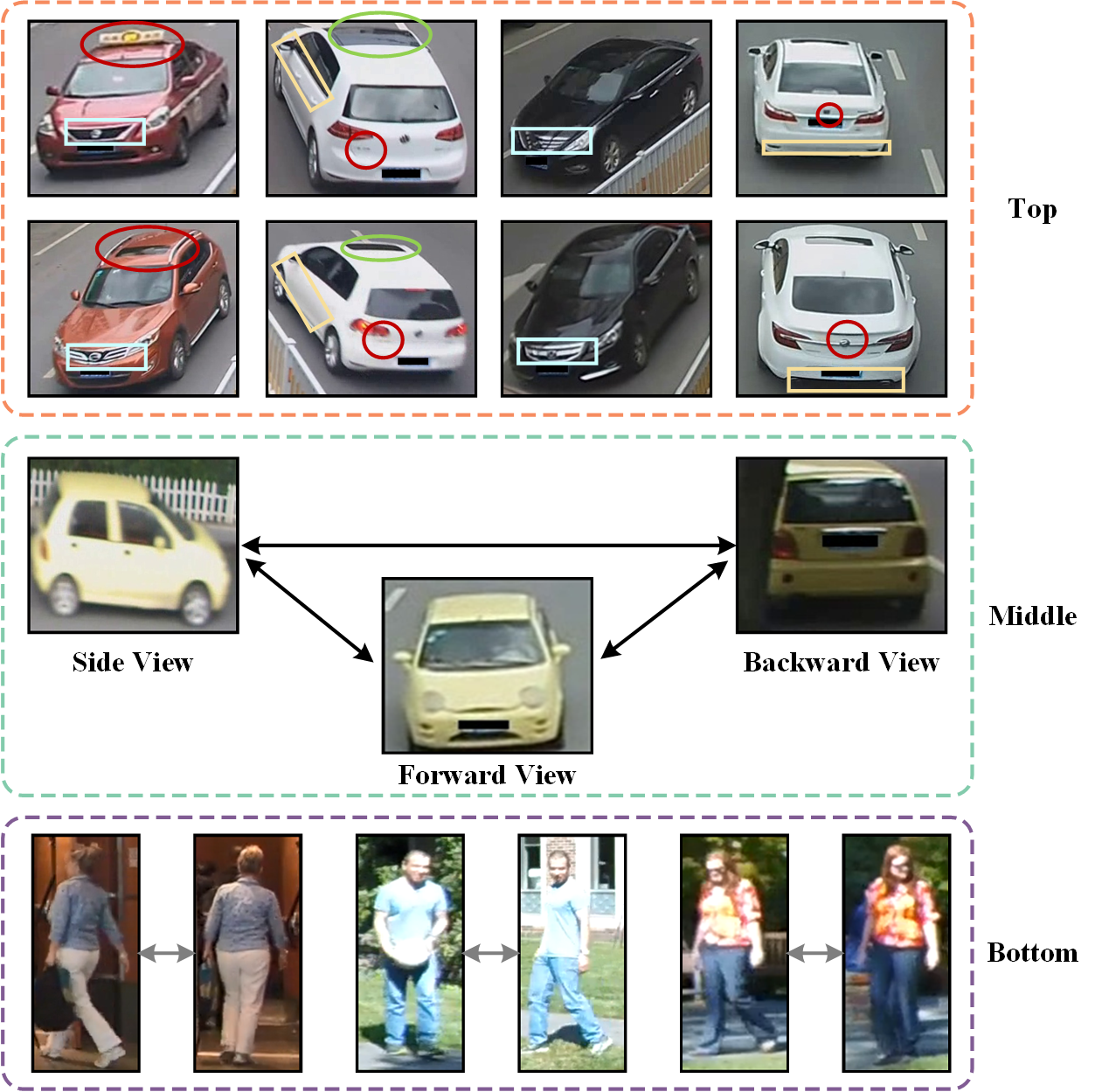}
		\setlength{\abovecaptionskip}{-3pt}
	\caption{
		(1) Different vehicles share extremely analogous appearances. 
		(2) Large intra-class variations. 
		(3) In person re-ID, the same person's images from different aspects share more visual overlap than those in vehicle re-ID. 
	}
	\label{fig1-challenges}
	\vspace{-15pt}
\end{figure}

A similar topic is person re-identification (re-ID), but it is totally different from vehicle re-ID in terms of the facing challenges.  
Compared to person re-ID, vehicle re-ID suffers from much smaller inter-class variations and larger intra-class variations, as shown in Fig. \ref{fig1-challenges}. 
In the top rectangle, each column represents a near-duplicate pair of vehicles with different identities. Due to diversified viewpoints, illuminations, and orientations of vehicle images, they only differ in slight as highlighted. 
Concretely, vehicles with distinct labels can be of the same model and the same color. Two different vehicles of the same model and color only differ in minor details, e.g. brands, individual decorations and scratches.  
Additionally, the same vehicle's images with various orientations, such as front, side, and rear, only share little visual overlap. As human sizes are smaller than vehicles, the images of one specific person with diverse viewpoints retain more appearance similarities. 

Considering the sophisticated relationship between intra-class and inter-class discrepancies, 
conventional methods typically focus on hand-craft features \cite{liao2015person, zheng2015scalable, zapletal2016vehicle}, including geometric attributes, color histograms, and textures. Their major disadvantage lies to the insensitivity to background clutter and large variations of light conditions and viewpoints, thus leading to restricted applications in practical situations. 
Recently, deep features have dominated many computer vision tasks, such as object detection \cite{chen2020reverse, feng2020residual, lee2020mercnn, liu2020deepsalient}, semantic segmentation \cite{pan2020joint} and action recognition \cite{dhiman2020view, ji2020deep, yang2020stacnn}.
Hence, researchers have embraced deep features into vehicle re-ID to construct a more efficient feature representation based on two strategies: 1) visual representation and 2) multi-modal representation. 

The first strategy exploits the visual representation of vehicles. Refs. \cite{liu2016deep, liu2016largescale, liu2016provid} devote to establish a global description of vehicle images. 
 However, the global features have limited performance when dealing with inevitable visual ambiguities among different vehicles and dramatic changes of uncontrolled variations of the same vehicle. This inspired 
several works \cite{Wang2017Orientation, he2019part, chen2019partition} to seek helpful visual clues for distinguishing subtle differences between vehicle images. 
However, these methods mainly focus on informative region localization instead of how local regions are assigned the importances to distinct degrees. 

To solve this problem, we develop a discriminative feature representation method to explore more fine-grained features by introducing an appearance module with two streams, 
i.e., the coarse-grained and the fine-grained feature streams respectively, to describe visual information of different granularities. The coarse-grained feature stream extracts deep features from the global network, presenting a macroscopic impression of images. Besides, the fine-grained feature stream pulls the samples of the same class up together while pushing those of different classes away in the feature embedding. 

The second strategy aims to discover and utilize multi-modal information to improve the performance of vehicle re-ID algorithms. Because visual appearance is not always reliable, especially in unconstrained environments with excessively dynamical changes,
including license plates \cite{liu2016deep} and vehicle models \cite{liu2016deep, guo2018learning, wei2018coarse, Rajamanoharan2019multitask}. 
As aforementioned, license plate recognition is vulnerable and involves privacy problems. Moreover, vehicle models require manually annotated labels, which are laborious and uneconomical. Contrastly, the spatio-temporal information of vehicle re-ID problem is usually available due to the universalness of video monitoring systems. 
Refs. \cite{liu2016provid, shen2017learning, Wang2017Orientation, lv2019vehicle, tan2019multi} employ the spatio-temporal information as the refinement of appearance features. Ref. \cite{liu2016provid} defines a spatio-temporal similarity between image pairs based on the approximate statistics of datasets. Ref. \cite{shen2017learning} uses Chain of Markov Random Field to model visual spatio-temporal paths. However, the missed detections of vehicles on passed spatio-temporal paths would degrade the vehicle re-ID algorithm's overall performance. Moreover, ref. \cite{lv2019vehicle} constructs spatio-temporal constraints and refines the matching problem by a transfer time matrix. However, the acquisition of timestamps in \cite{lv2019vehicle} requires the preprocess of a multi-camera multi-target tracking task, which injures the method's adaptability and extendibility.
Additionally, a group of works for person re-ID also explore spatio-temporal information, which can be mainly divided into two manners. One is to excavate implicit spatial-temporal information in videos \cite{ wang2018p2snet, li2018diversity}. The other is to use explicit spatio-temporal information as physical constraints to reduce the complexity of the matching algorithm \cite{cho2019joint, lv2018unsupervised}. 
The spatio-temporal model for person re-ID can not be straightforwardly used to vehicle re-ID due to severe performance degradation. 
In particular, since vehicles move much faster than people, the assumption of constrained location prediction \cite{lv2018unsupervised} is not rational any more.

To solve the above problem, we propose a spatio-temporal module to obtain a more robust model for identifying vehicles. Specifically, the distances between camera pairs and the time intervals are modeled as a distribution rather than a transfer matrix \cite{lv2019vehicle}, spatio-temporal constraints \cite{tan2019multi} or visual spatio-temporal paths \cite{shen2017learning}. By modeling the camera locations' distance and the discrepancy of timestamps as random variables, we formulate the spatio-temporal relationship in a simple yet effective manner quantitively. By adding the spatio-temporal module, we observe an evident performance improvement. 

In summary, we propose a novel discriminative representation with spatio-temporal information (DFR-ST) to establish a robust feature embedding with multi-modal cues for vehicle re-ID.
The main contributions of DFR-ST are three-fold: 
\begin{itemize}
	\item The proposed DFR-ST constructs the appearance representation by the two-stream architecture to extract the coarse-grained and the fine-grained features. Besides, the combination of an attention mechanism and division operations drives the fine-grained visual representation to focus on more salient and informative regions.
	\item The spatio-temporal module is proposed to form a complementary representation with the visual appearance by taking multi-modal cues into sufficient considerations.  
	\item Extensive experiments on two large-scale benchmarks indicate the effectiveness and robustness of the proposed DFR-ST and it achieves the state-of-the-art performance. 
\end{itemize}

The rest of this article is organized as follows. 
Section \ref{section:related works} refers to the related works of vehicle re-ID and person re-ID. 
Section \ref{section:3} introduces the proposed DFR-ST method while Section \ref{section:experiments} presents the experimental results and analyses. Section \ref{section:conclusion} concludes this article.

\section{Related Works}
\label{section:related works}

\subsection{Vehicle Re-Identification}
The earliest vision-based works \cite{liao2015person, zheng2015scalable, zapletal2016vehicle} design hand-craft features scrupulously to identify vehicles. 
However, hand-craft features have limited capability in practice, because heavy occlusions and drastic light changes in unconstrained situations would destroy the performance when modeling discriminative features.
Since deep features exhibit powerful strength on multiple vision tasks including  image classification \cite{chang2020devil, meng2020constrained, sun2020supervides}, object detection \cite{chen2020reverse, , feng2020residual, lee2020mercnn, liu2020deepsalient} and action recognition \cite{dhiman2020view, ji2020deep, yang2020stacnn}, researchers exploit deep features for vehicle re-ID mainly through two approaches as follows. 

The first approach is constructing visual representation to tackle the issue of identifying the same vehicle. 
Wang \emph{et al.} \cite{Wang2017Orientation} aggregate four local features with different directions to describe an orientation-invariant feature. However, this work requires a manual classification of the key points on vehicles.
He \emph{et al.} \cite{he2019part} utilize an object detection method to extract partial features for near-duplicate vehicles. 
Chen \emph{et al.} \cite{chen2019partition} propose a two-branch network with partitions on the height and the width channels to maximally distinct local features. The experimental results illustrate that the channel-wise partition improves most. 
Our work shares a similar idea with PRN \cite{chen2019partition}, which fuses the global and the local features to establish an overall embedding. However, observing that the above methods focus on the local region localization instead of assigning different degrees of importance on different informative regions, we design a more discriminative representation by an attention network and the collaboration of three-dimension divisions. Moreover, experiments demonstrate that our algorithm performs better than \cite{chen2019partition} on public datasets.

The second approach aims to improve vehicle re-ID performance with multi-modal information. The most intuitive idea is the introduction of license plates \cite{liu2016deep}. However, this unique information relies on the performance of character recognition tasks, which inevitably suffer from low-resolution videos, frequent occlusions and blurred motion. Besides, the model type information of vehicles is also be explored to benefit matching vehicles. Guo \emph{et al.} \cite{guo2018learning} propose a structured feature embedding by defining a vehicle model classification loss and two ranking losses with distinct granularities. Although the fine-grained ranking loss has a huge contribution to the final performance, acquiring vehicle models with manually annotation is uneconomical and laborious. 

Contrastly, spatio-temporal information for vehicle re-identification is usually available because of the rapid popularization of video monitoring systems. Hence, it is more efficient to incorporate spatio-temporal cues in the algorithms to complement the visual appearance information with lower costs. 
Wang \emph{et al.} \cite{Wang2017Orientation} propose the orientation-invariant visual representation to describe the macroscopic embedding of vehicles with constraints on spatio-temporal relationships. 
Shen \emph{et al.} \cite{shen2017learning} establish candidate paths using Chain of Markov Random Field and they employ a siamese architecture with long short term memory (LSTM) network units to model the visual appearance and spatio-temporal information. 
Lv \emph{et al.} \cite{lv2019vehicle} adopt the combination of three different features learned by various losses to identify vehicles. The spatio-temporal constraints is realized by a transfer time matrix, which refine the search space and reduce the computing complexity.
However, we notice that the existing models of spatio-temporal information are relatively qualitative and intuitive, which is lack of a more precise mathematical description. Therefore, we propose a spatio-temporal module to construct multi-modal representation by modeling camera locations and time intervals as random variables. 
Different from \cite{Wang2017Orientation}, which only focuses on transition time intervals but neglects spatial information, we consider both temporal and spatial clues simultaneously by a quantitative formulation. Experiments demonstrate the effectiveness of the proposed spatio-temporal module.

\subsection{Person Re-Identification}
 Most existing vision-based approaches could be classified into two categories. 
 The first category focuses on improving feature representation against pose variants, clutter backgrounds, and distinct camera viewpoints. The most common strategy is to train the deep network on multiple local regions and combine all local branches with the global one \cite{luo2019alignedreid, wang2020cdpm, zhang2020multiscale}. 
 The second category considers metric learning \cite{schroff2015facenet, zhu2018video, zhu2020finegrained}. 
Besides, some works utilize multi-modal information including RGB-D data \cite{ren2019uniform}, pose estimation \cite{cho2018pamm} and segmentation annotations \cite{huang2020improve}, because the collaboration of multi-modal data can obtain further performance boosting. However, the cost of data acquisition is an unnegligible problem. As for vehicle re-ID, camera locations and timestamps of videos are easy to obtain because of surveillance systems' prevalence. Hence, taking spatio-temporal information into considerations is a reasonable proposal and can promote the performance of vehicle re-ID methods without much higher costs. 
Note that our proposed spatio-temporal module does not need camera calibration information, which is usually unavailable and contains additional computing costs and noises.

\begin{figure}[!t]
	\centering
	\includegraphics[width=3in]{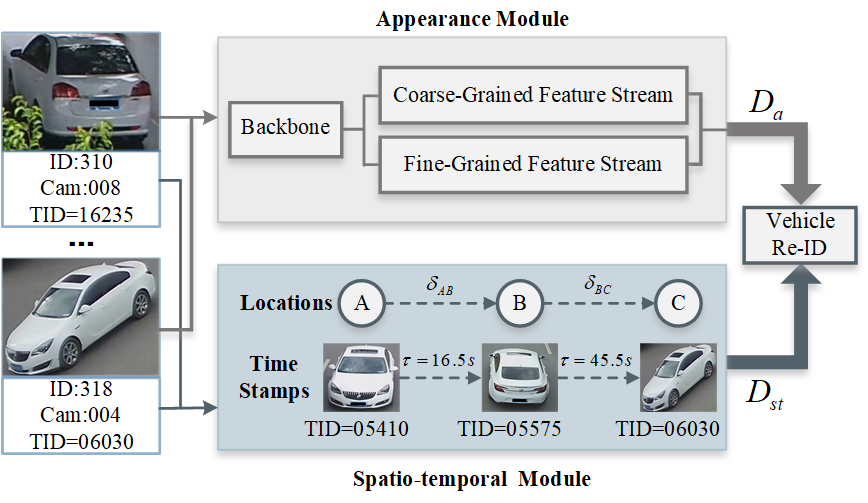}
		\setlength{\abovecaptionskip}{-3pt}
	\caption{An overview of our DFR-ST approach. The proposed method contains the appearance module and the spatio-temporal module—the former aims to establish a discriminative feature embedding through a two-stream structure by involving multi-grained features, and the latter uses camera locations and timestamps as the spatio-temporal cues to construct further refinement. 
	}
	\label{fig2_framework}
	\vspace{-15pt}
\end{figure}

\section{Methodology}
\label{section:3}
In this section, we design an appearance module and a spatio-temporal module, 
aiming to empower the capability of DFR-ST for discriminative and robust feature representation. Detailed descriptions of each module in the proposed DFR-ST method is demonstrated in the following sub sections.

\subsection{System Architecture}
\begin{figure*}[!t]
	\centering
	\includegraphics[width=6.5in]{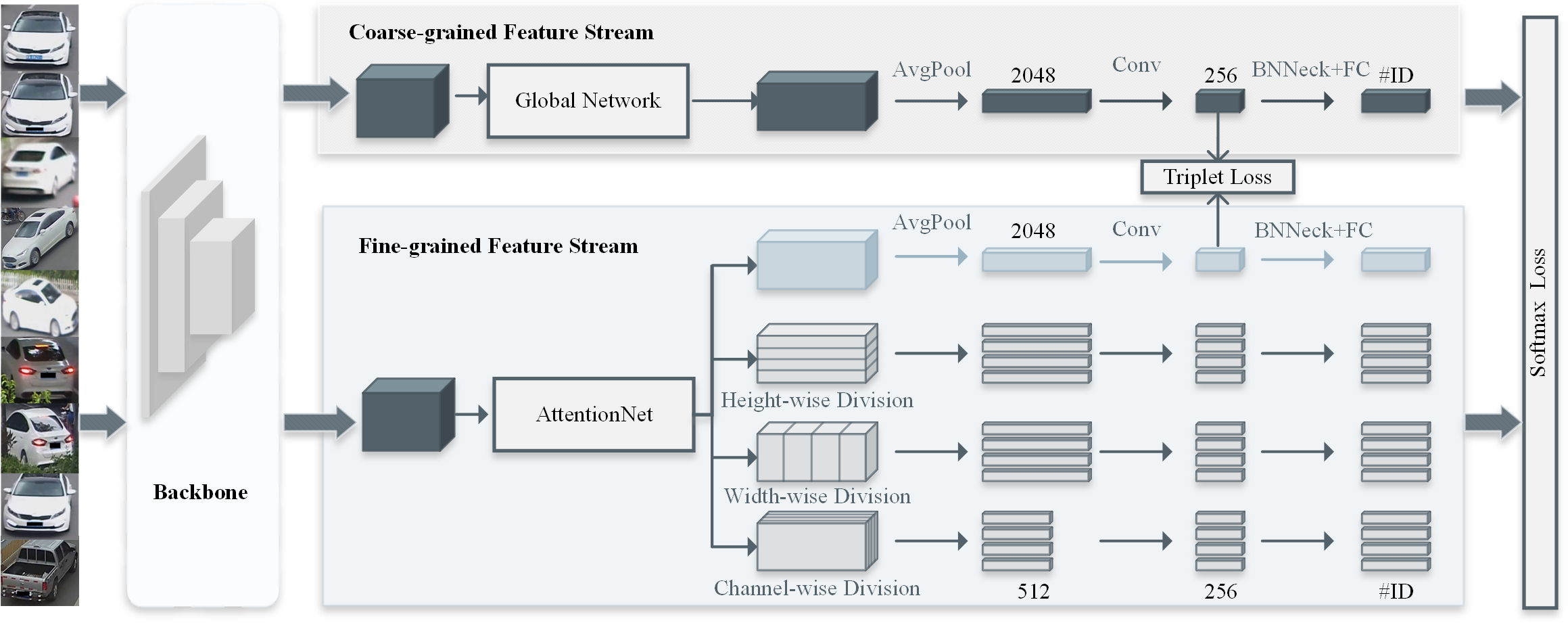}
		\setlength{\abovecaptionskip}{-3pt}
	\caption{The schematic diagram of the appearance module in DFR-ST. The shallow representation extracted from the backbone is transmitted to the coarse-grained feature stream for constructing a macroscopic feature embedding and the fine-grained feature stream for a microscopic representation simultaneously. The fine-grained feature stream contains an AttentionNet and four branches including division operations of three dimensions to extract high-quality part-level representations. Afterward, the aggregation of two-stream features constructs a discriminative feature representation for identifying the same vehicle. 
	}
	\label{fig8-appeanrance}
	\vspace{-15pt}
\end{figure*}

Fig. \ref{fig2_framework} illustrates the overall framework of the proposed DFR-ST. The appearance module is responsible for modeling visual features, in which input images are fed into a backbone network followed by two streams: \emph{Coarse-Grained Feature Stream} and \emph{Fine-Grained Feature Stream}, as shown in Fig. \ref{fig8-appeanrance}. 
Meanwhile, the spatio-temporal module establishes the spatial and temporal distances to provide extra information for identifying the same vehicle. Finally, the combination of visual and spatio-temporal representation measures the similarity of vehicle images, conducting the ranking list of gallery images.

The appearance module is composed of a coarse-grained feature stream and a fine-grained feature stream. The coarse-grained feature stream extracts the general feature representation $\mathbf{x}_{c}$ to deal with the complicated relationship of the inter-class and intra-class variation. This stream attempts to enlarge the distances of the samples with distinct identities in the embedding space. However, only general features could fail to process the detailed discrepancies of input images, especially dealing with the images with only slight differences such as private decorations, irregular scratches, and brands. Hence, $\mathbf{x}_{c}$ is replenished with $\mathbf{x}_f$ from the fine-grained feature stream, which can take account of local regions and salient parts. 

Moreover, we propose a spatio-temporal module to exploit extra minutiae, considering the images with intricate changes in unconstrained environments degrade the visual appearance's effectiveness in practical scenarios, 
In particular, we apply the camera location and the timestamps information as the spatio-temporal cues, which are usually easily acquired due to the prevalence of the security video systems. 
Thus, the collaboration of the visual appearance representation and the spatio-temporal clues enhances the final vehicle re-ID performance.

\subsection{Appearance Module}

\subsubsection{Coarse-Grained Feature Stream}
The coarse-grained feature stream extracts a macroscopic description of input images. Features obtained from the backbone are delivered to the coarse-grained feature stream for further processing. Two residual blocks are placed successively as the Global Network, followed by an average pooling operation and a $1 \times 1$ convolution operation. To promote the overall performance, we adopt the BNNeck strategy \cite{luo2019bag} to alleviate the inconsistency of the triplet loss and the ID loss in the embedding. Moreover, we set the stride of down-sampling operations in the last convolutional layer 
to 1 to maintain more deep information.

\subsubsection{Fine-Grained Feature Stream}
Opposed to the coarse-grained stream, the fine-grained feature stream captures the local regions' microcosmic features to distinct challenging near-duplicate vehicles. The AttentionNet receives the backbone network features to construct an attentive feature representation, followed by four branches for further processing. The first branch considers assigning reasonable weights to different local regions. And the rest three branches conduct the divisions of features along three dimensions, i.e., the height-wise, the width-wise, and the channel-wise operations. 
The division strategy is inspired by the height slicing strategy in the person re-ID approach \cite{luo2020alignedreid}. But unlike the vertical partition of a human body, a single-vehicle has semantic partitions along both vertical and horizontal dimensions. Specifically, the vehicle is composed of the car roof, the window, the bumper, the license plate, the chassis in the vertical dimension, and the rearview mirror, the door, the main body in the horizontal dimension. Hence, we design the channel, the height, and the width division branches concurrently because different channels carry independent semantic information. 

\begin{figure}[!t]
	\centering
	\includegraphics[width=3.5in]{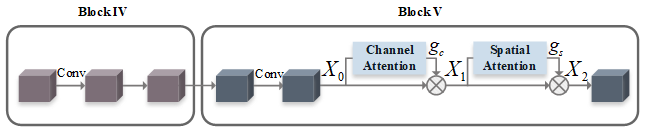}
		\setlength{\abovecaptionskip}{-10pt}
	\caption{The architecture of AttentionNet. The sequential placement of the channel and spatial attention are added in the network's last residual block to further learn an attentive feature representation from local regions adaptively.}
	\label{fig-resblockAttention}
	\vspace{-10pt}
\end{figure}

The architecture of AttentionNet is shown in Fig. \ref{fig-resblockAttention}. The feature $\mathbf{X}_0\in\mathbb{R}^{C\times H\times W}$ is firstly processed with the channel-domain attention to learn a channel feature map $\mathbf{g}_c\in\mathbb{R} ^{C}$, then followed by the spatial-domain attention with the learned attention map $\mathbf{g}_s\in\mathbb{R} ^{1\times H\times W}.$ $C, H, W$ denote the channel, height, and width dimensions of $\mathbf{X}_0$. The above procedure is:
\begin{align*}
{\setlength\abovedisplayskip{1pt plus 3pt minus 7pt}
	\setlength\belowdisplayskip{1pt plus 3pt minus 7pt}}
\mathbf{X}_1 (i, j, k) &= \mathbf{g}_c (i)  \mathbf{X}_0 (i, j, k), \forall i, \\
\mathbf{X}_2 (i,j,k) &= \mathbf{g}_s (j, k) \mathbf{X}_1 (i,j,k), \forall j, k.
\end{align*}
In particular, $\mathbf{X}_1$ and $\mathbf{X}_2$ denote the output features after channel and spatial attention operations successively. $i,j,k$ are indexes of the channel, height, and width dimensions.

\begin{figure}[!t]
	\centering
	\includegraphics[width=3.5in]{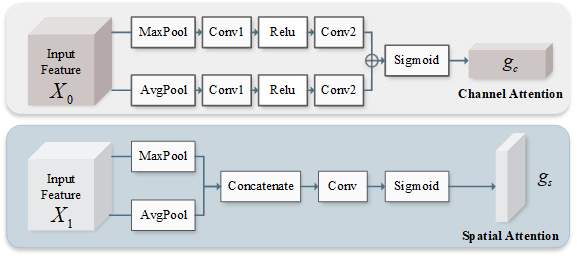}
		\setlength{\abovecaptionskip}{-10pt}
	\caption{The block diagrams of the channel and spatial attention operations. $\oplus$ denotes the element-wise summation. $\mathbf{g}_c$, $\mathbf{g}_s$ denote the learned channel and spatial attention maps respectively.}
	\label{fig3-attention}
	\vspace{-15pt}
\end{figure}

Fig. \ref{fig3-attention} illustrates the designed structure of channel attention and spatial attention. 
Define the transformation $\mathcal{T}_c$: $\mathbf{X}_0\rightarrow\mathbf{g}_c$.
The input feature $\mathbf{X}_0$ is first transmitted to the squeeze operation constituted of a global average pooling and a maximum pooling. The parallel pooling operations can aggregate feature maps along the spatial dimension, symbolizing the overall distribution of channel-domain responses. Second, a multi-layer perceptron machine with one hidden layer processes the element-wise summation of two pooling operations. Finally, a sigmoid function is applied to grasp channel-domain dependencies. The above procedure is:
\begin{align*}
{\setlength\abovedisplayskip{1pt plus 3pt minus 7pt}
	\setlength\belowdisplayskip{1pt plus 3pt minus 7pt}}
\mathbf{g}_c&=\mathcal{T}_c(\mathbf{X}_0)	     =\sigma\{\text{MLP}[\text{P}_{\text{avg}}(\mathbf{X}_0)] +\text{MLP}[\text{P}_{\text{max}}(\mathbf{X}_0)]\}\\
&=\sigma \{\mathbf{W}_2 \delta [\mathbf{W}_1 (\mathbf{x}^\text{c}_{\text{avg}})] +  \mathbf{W}_2 \delta [\mathbf{W}_1 (\mathbf{x}^\text{c}_{\text{max}})] \},
\end{align*}
where
\begin{align*}
{\setlength\abovedisplayskip{1pt plus 3pt minus 7pt}
	\setlength\belowdisplayskip{1pt plus 3pt minus 7pt}}
\mathbf{x}^\text{c}_{\text{avg}} (i) &= \frac{1}{W\times H} \sum^W_{j=1} \sum ^H_{k=1} \mathbf{X}_0(i,j,k),\\
\mathbf{x}^\text{c}_{\text{max}} (i) &= \max_{j,k} \mathbf{X}_0(i,j,k).
\end{align*}
In particular, $\sigma$ is a sigmoid function and $\delta$ represents a ReLu function. $\text{MLP}$ denotes a multi-layer perception machine with one hidden layer. 
$\text{P}_{\text{avg}}$ and $\text{P}_{\text{max}}$ are average pooling and maximum pooling operators. 
$\mathbf{W}_1\in\mathbb{R}^{\frac{C}{k}\times C}$ and $\mathbf{W}_2\in\mathbb{R}^{C\times\frac{C}{k}}$ refer to the weights of two convolution layers in $\text{MLP}$, where the reduction ratio $k$ is 16 in the experiments. 
$\mathbf{x}^\text{c}_{\text{avg}}$ and $\mathbf{x}^\text{c}_{\text{max}}$ are the average and maximum pooling results of $\mathbf{X}_0$.

Spatial attention concentrates on the location information of salient parts. Define the transformation $\mathcal{T}_s:\mathbf{X}_1\rightarrow\mathbf{g}_s$.
The squeeze operation processes $\mathbf{X}_1$ with average pooling and maximum pooling operations across the channel dimension. Moreover, the concatenation of pooled feature maps is passed through a single convolution layer, followed by a sigmoid function. 
Mathematically, 
\begin{equation*}
\mathbf{g}_s=\mathcal{T}_s(\mathbf{X}_1)
=\sigma\{\mathbf{W} ( [\mathbf{x}^\text{s}_{\text{max}}; \mathbf{x}^\text{s}_{\text{avg}}]) \},	
\end{equation*}
where
\begin{align*}
\mathbf{x}^\text{s}_{\text{max}} (j,k) &= \max_{i} \mathbf{X}_1(i,j,k), \forall j,k,\\
\mathbf{x}^\text{s}_{\text{avg}} (j,k) &= \frac{1}{C}\sum^C_{i=1} \mathbf{X_1} (i,j,k), \forall j,k.
\end{align*}
Specifically, $\sigma$ denotes the sigmoid function, and $\mathbf{W}$ is the weight of the convolution layer. $\mathbf{x}^\text{s}_{\text{max}}$ and $\mathbf{x}^\text{s}_{\text{avg}}$ represent the output feature maps of maximum pooling and average pooling operations across the channel axis. 

\subsubsection{Object Function}
The total loss is the weighted summation of a cross-entropy loss and a triplet loss. The loss of a batch with $N$ images is written as: 
\begin{equation}
\label{eq:loss}
L_{\text{all}} = L_{\text{ce}} + \lambda L_{\text{tri}},
\end{equation}
where $L_{\text{ce}}$ is the cross-entropy loss and $L_{\text{tri}}$ is the triplet loss with the batch hard sampling strategy. $\lambda$ is a weitht hyperparameter. 
The label smoothing strategy \cite{Szegedy2015rethinking} is employed with the cross-entropy loss to alleviate the overfitting problem. Hence, the cross-entropy loss $L_{\text{ce}}$ can be delineated as:
\begin{equation}
\label{eq:crossentropy}
L_{\text{ce}} = - \frac{1}{N} \sum_{i=0}^{N-1} \sum_{k=0}^{K-1} p_{i,k} \log q_{i,k}
\end{equation}
where
\begin{align*}
p_{i,k}=\left\{
\begin{aligned}
&1-\varepsilon, &\text{if} \   y_i=k \\
&\frac{\varepsilon}{K-1}, &\text{if} \   y_i \neq k
\end{aligned}
\right.,
\quad q_{i,k}=\frac{\exp[\Phi(\mathbf{I}_i)]}{\sum_{k=0}^{K-1} \exp[\Phi (\mathbf{I}_k)]}.
\end{align*}
In particular, $i\in\{0,\cdots,N-1\}$ is the index of images in the batch and $k\in\{0,\cdots,K-1\}$ is the index of  $K$ classes. $p_{i,k}$ denotes the distribution after label smoothing. $y_i$ is the ground truth label of the $i$th image $\mathbf{I}_i$. 
The hyperparameter $\varepsilon \in [0,1]$ is a weight factor. 
$q_{i,k}$ is the network prediction probability of $\mathbf{I}_i$ to the $k$th class and $\Phi(\cdot)$ means the transformation of the appearance module. 
Furthermore, the triplet loss $L_{\text{tri}}$ is:
\begin{equation}
L_{\text{tri}} = \sum_i^N [||\mathbf{x}_i^a-\mathbf{x}_i^p||_2^2 -
||\mathbf{x}_i^a-\mathbf{x}_i^n ||_2^2+m ]_+,
\label{eq:tripletloss}
\end{equation}
where $\mathbf{x}_i^a$, $\mathbf{x}_i^p$, $\mathbf{x}_i^n$ denote features of anchors, positive and negative samples respectively. $[\cdot]_+$ represents the hinge function, and $m$ controls the margin of distances between the positive and the negative samples to the anchors.

\subsubsection{Backbone}
The backbone network establishes a shallow representation of visual appearance. 
In the proposed DFR-ST, the backbone network adopts the first three blocks of ResNet-50 network for its flexible architecture and superior performance. As shown in Fig. \ref{fig8-appeanrance}, we duplicate subsequent convolutional layers after the first three residual blocks to split ResNet-50 network into two feature branches. Different convolutional network architectures designed for deep learning also can be adjusted as the backbone correspondingly.

\subsection{Spatio-temporal Module}
In real-world applications, the appearance model is not sufficient to construct a discriminative representation in the presence of noises from intricate brackgrounds and severe occlusions. Therefore, camera location and timestamp information, available in urban surveillance and intelligent transportation systems, can provide extra content on vehicle attributes. 

The fundamental assumption is that two images with smaller spatial or temporal distances have higher possibilities to be the same identity based on the observation in \cite{liu2016deep}, and vice versa. According to this assumption, we propose a novel spatio-temporal measurement, in which the spatial similarity $D_s$ and the temporal similarity $ D_t$ between a query image $i$ from camera $c_i$ and a gallery image $j$ from camera $c_j$ are as follows:
\begin{align}
\label{eq:beta}
D_s &= \frac{1}{1+\exp (\alpha_1 [ p(\delta |\mu_\delta, \sigma_\delta) -  \alpha_2])}, \\
D_t &= \frac{1}{1+\exp(\beta_1 [ p(\tau | \mu_\tau, \sigma_\tau) - \beta_2 ])},
\label{eq:alpha}
\end{align}
where $\delta$ denotes the shortest distance on Google map between camera $c_i$ and $c_j$. $\tau$ is the discrepancy of time stamps between $i$ and $j$. The hyperparameters $\alpha_1$, $\alpha_2$ and $\beta_1$, $\beta_2$ indicate that  higher probilities corresponding to smaller spatio-temporal distances. $p(\delta |\mu_\delta, \sigma_\delta)$ and $p(\tau | \mu_\tau, \sigma_\tau)$ describe the estimation of conditional probabilities of $\delta$ and $\tau$ with parameters $(\mu_\delta, \sigma_\delta)$ and $(\mu_\tau, \sigma_\tau)$, which are modeled as log-normal distributions:
\begin{align*}
p(\delta |\mu_\delta, \sigma_\delta) &= 
\ln \mathcal{N}(\delta;\mu_\delta, \sigma_\delta) =
\frac{1}{\delta \sqrt{2 \pi \sigma_\delta^2}} \exp[-\frac{(\ln \delta-\mu_\delta )^2}{2\sigma_\delta^2} ], \\
p(\tau | \mu_\tau, \sigma_\tau) &= 
\ln \mathcal{N}(\tau;\mu_\tau, \sigma_\tau) =
\frac{1}{ \tau \sqrt{2\pi \sigma_\tau^2}} \exp[-\frac{(\ln\tau-\mu_\tau )^2}{2\sigma_\tau^2} ].
\end{align*}
The parameters $(\mu_\delta, \sigma_\delta)$ and $(\mu_\tau, \sigma_\tau)$ of the distributions are estimated by maximizing the following likelihood functions:
\begin{align*}
L(\delta|\mu_\delta, \sigma_\delta)&= \prod_{i=1}^N (\frac{1}{\delta_i})\mathcal{N}(\ln \delta_i;\mu_\delta, \sigma_\delta), \\
L(\tau|\mu_\tau, \sigma_\tau) &= \prod_{i=1}^N (\frac{1}{\tau_i}) \mathcal{N}(\ln \tau_i;\mu_\tau, \sigma_\tau).
\end{align*}
where $N$ is the number of images in the gallery set.

In this way, we establish the spatio-temporal module to provide an additional refinement for vehicle re-ID. Among the existing spatio-temporal models \cite{liu2016deep, Wang2017Orientation, shen2017learning, tan2019multi, lv2019vehicle}, the most similar spatio-temporal model is \cite{Wang2017Orientation}. 
Different from \cite{Wang2017Orientation}, which considers the transition time intervals of vehicles but does not model the spatial information, our DFR-ST involves both temporal and spatial cues simultaneously. 
Therefore, we can explicitly obtain a quantitative formulation of spatio-temporal relationships to promote overall performance. 


	Hence, the entire retrieval process is:  DFR-ST firstly extracts an appearance representation $D_a$ from the appearance module, which describes the distance of the query image $i$ and the gallery set in the feature embedding space. 
	Secondly, the spatio-temporal module constructs the spatio-temporal similarity $D_{st}$ to evaluate the similarities in spatial and temporal domains.
	The overall similarity between the query image $i$ and the gallery image $j$ is calculated as:
	\begin{equation}
	D(j) = D_a(j) + \omega [D_s(i,j) +D_t(i,j)], \forall i.
	\label{eq:distance}
	\end{equation}
	$\omega$ is a weighted parameter. $D_s(i,j)$ and $D_t(i,j)$ model the spatial and temporal similarities between $i$ and $j$ respectively. 
	Finally, the ranking list of the proposed DFR-ST is conducted by the weighted summation of $D_a$ and $D_{st}$. 

\section{Experiments}
\label{section:experiments}
In this section, the proposed DFR-ST is evaluated on public datasets and compared with state-of-the-art methods. Extensive experiments are conducted to demonstrate the effectiveness and the robustness of DFR-ST at the end of this section.

\subsection{Datasets}
The experiments are executed on two public large-scale datasets, \emph{i.e.}, VeRi-776 \cite{liu2016deep} and VehicleID \cite{liu2016relative}. The following subsections introduce the two datasets and evaluation metrics.

\begin{figure*}[!t]
	\centering
	\includegraphics[width=6in]{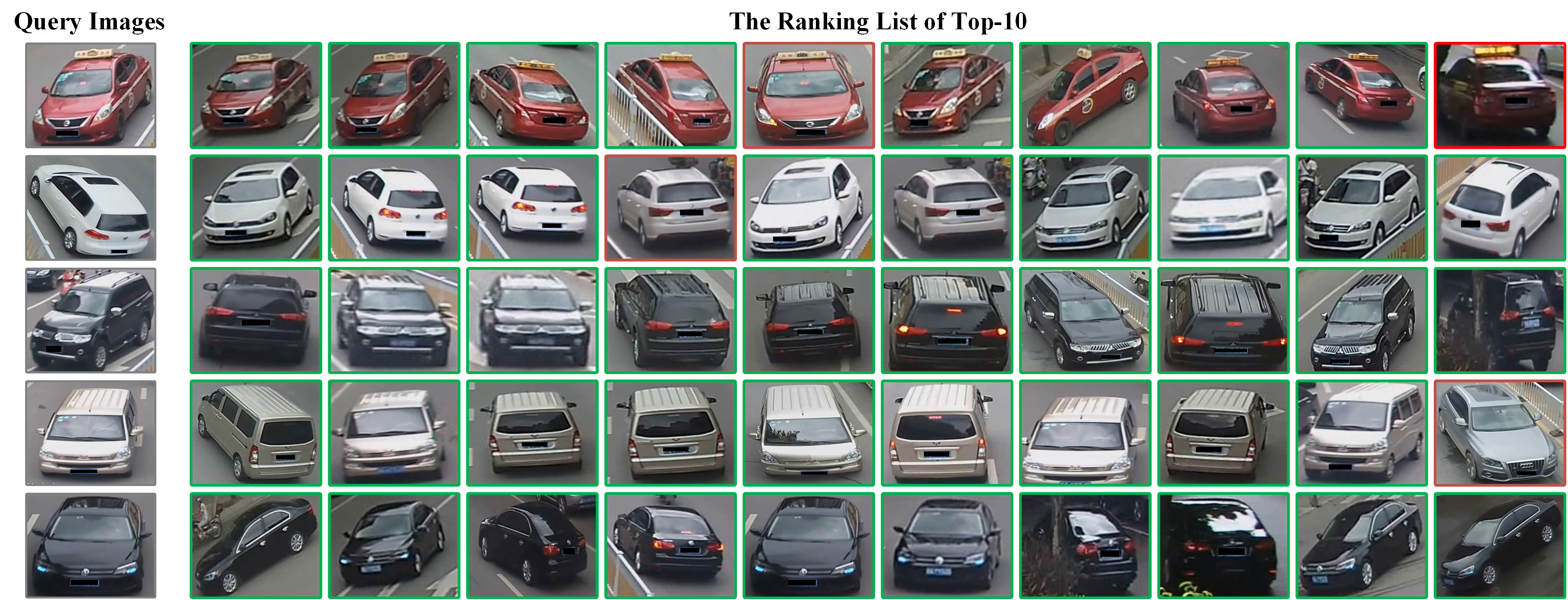}
			\setlength{\abovecaptionskip}{-3pt}
	\caption{Visualization of the ranking lists of DFR-ST on VeRi-776. The re-identification results are listed in ascending order of the appearance and the spatio-temporal distance between the query image and the gallery images. The green (red) boxes denote the correct (wrong) re-identification. 
	}
	\label{fig4-veri-visualizer}
	\vspace{-15pt}
\end{figure*}

\subsubsection{VeRi-776}
The VeRi-776 dataset contains 49,357 images of 776 different vehicles captured by 20 cameras involving various viewpoints, illumination changes, and background clutters. The training set consists of 37,778 images with 576 identities, and the test set receives the remaining 11,579 images with 200 identities. Moreover, the selected 1,678 images of the testing set constitute the query set.
Followed \cite{liu2016deep}, the evaluation metrics of VeRi-776 are mean Average Precision (mAP), Top-1, and Top-5 accuracy of Cumulative Match Curve (CMC) corresponding to the image-to-track search.

\subsubsection{VehicleID}
VehicleID dataset is released after the VeRi-776, which includes 221,763 images with 26,267 identities and 250 models. Researchers annotate 90,196 images with model labels. Among the identities, the training set contains 13,164 identities, and the rest remains for the test set. Three test splits for different gallery sizes are 800, 1,600 and 2,400, as depicted in Table \ref{table1:test splits}. 
For each test split, the test set randomly selects images of distinct identities to form the gallery set, and the same procedure repeats ten times. The average results of 10-times procedures are the final performance. Evaluation metrics on VehicleID are mAP, Top-1, and Top-5 accuracy of CMC.

	\begin{table}[!t]
	\renewcommand{\arraystretch}{1.3}
	\caption{Three test splits of VehicleID dataset.}
	\label{table1:test splits}
	\centering
	\begin{tabular}{cccc}
		\toprule
		& Small &Medium & Large\\
		\midrule
		Query Images &800 & 1,600 &2,400\\
		Gallery Images & 6,493 & 13,377 & 19,777\\
		\bottomrule
	\end{tabular}
	\vspace{-15pt}
\end{table}

\subsection{Implementation Details}
The backbone network for shallow feature extraction adopts ResNet-50 \cite{he2016deep}. In the coarse-grained feature stream, the average pooling operation is employed after three residual blocks and a $1 \times 1$ convolutional layer is followed. $\lambda$ in Eq. (\ref{eq:loss}) is set to 0.4 and $\varepsilon$ in Eq. (\ref{eq:crossentropy}) is set to 0.1. $m$  in Eq. \ref{eq:tripletloss} is set to 1.2.
Moreover, the parameters $\alpha_1$ and $\alpha_2$ are set to 6 while $\beta_1$ and $\beta_2$ are set to 0.5 in Eq. (\ref{eq:beta}) and Eq. (\ref{eq:alpha}). The weighted parameter $\omega$ in Eq. (\ref{eq:distance}) is set to 0.2.

As for the training procedure, we apply the adam optimizer with a weight decay $5\emph{e}^{-4}$ and a synchronous batch normalization strategy. The initial learning rate is $1\emph{e}^{-4}$, which is adjusted by a warmup strategy.  Moreover, we utilize the Euclidean distance to compute the similarity between the query and gallery images during training and testing. The batch size is 32, with randomly selected eight identities and four images of each identity. We train the appearance module of DFR-ST for 135 epochs on two Nvidia GeForce GTX 1080 Ti GPUs. The overall DFR-ST is implemented on the PyTorch platform.
Moreover, our work adopts the re-ranking strategy \cite{zhong2017reranking} for further boosting the performance. 
since re-ranking benefits performance promotion  \cite{li2019common, zhong2017reranking, guo2018learning, shankar2019comparative}.

\begin{table}[!t]
	\renewcommand{\arraystretch}{1.3}
		\setlength{\abovecaptionskip}{-3pt}
	\caption{Comparison with state-of-the-art on VeRi-776.}
	\label{table2:veri}
	\centering
	\begin{tabular}{cccc}
		\toprule
		Methods & mAP & Top-1 & Top-5\\
		\midrule
		LOMO \cite{liao2015person} & 9.64 & 25.33 & 46.48 \\ 
		BOW-CN \cite{zheng2015scalable}   & 12.20 & 33.91 & 53.69 \\ 
		FACT \cite{liu2016largescale} & 19.92 & 59.65 & 75.27\\  
		SCCN-Ft + CLBL-8-Ft \cite{zhou2018vehicle} & 25.12 & 60.83&78.55 \\ 
		OIN \cite{Wang2017Orientation} & 48.00& 65.9 & 87.7 \\ 
		VAMI \cite{zhou2018viewpoint} & 50.13 & 77.03 & 90.82 \\ 
		RNN-HA (ResNet) \cite{wei2018coarse} & 56.80 & 74.49 & 87.31 \\ 
		Hard-View-EALN \cite{lou2020embedding} & 57.44 & 84.39 & 94.05 \\ 
		GRF + GGL \cite{liu2020group} & 61.7 & 89.4 & 95.0\\ 
		QD-DLF \cite{zhu2020vehicle}      & 61.83 & 88.50 & 94.46 \\ 
		SPAN w/ CPDM \cite{chen2020orientationaware} &68.9 &94.0&97.6\\ 
		SAVER \cite{Khorramshahi2020devil} &79.6&96.4&98.6\\ 
		HPGN \cite{shen2020exploring} &80.18& 96.72& N/A\\ 
		PRN \cite{chen2019partition} &85.84&97.14& $\mathbf{99.40}$ \\
		PRN + ReRanking \cite{chen2019partition}    & \underline{90.48} & \underline{97.38} & 98.87 \\ 
		\hline
		FACT + Plate-SNN + STR \cite{liu2016deep} & 27.77 & 61.44 & 78.78\\ 
		PROVID \cite{liu2016provid} & 53.42 & 81.56 & 95.11 \\ 
		OIN + ST \cite{Wang2017Orientation} & 51.42 & 68.3 & 89.7 \\ 
		Siamese-CNN + Path-LSTM \cite{shen2017learning} & 58.27 & 83.49 & 90.04 \\ 
		VAMI \cite{zhou2018viewpoint} + STR \cite{liu2016largescale}  & 61.32 & 85.92 & 91.84 \\ 
		ReID + query expansion \cite{lv2019vehicle} & 70.8 & 93.2 & 98.0 \\ 
		\hline
		DFR-ST (ours)   &  86.00 & 95.67 & $\underline{99.17}$ \\
		DFR-ST (ours) + ReRanking  &  $\mathbf{91.56}$  & $\mathbf{97.74}$ & 98.41 \\
		\bottomrule
	\end{tabular}
	\vspace{-15pt}
\end{table}

\subsection{Performance Comparison on VeRi-776}
The proposed DFR-ST is compared with state-of-the-art methods on a large-scale public dataset, i.e., VeRi-776, and the results are presented in Table \ref{table2:veri}. Note that we 
directly copy state-of-the-art algorithms' performance from the original papers instead of reproducing all methods. 

On VeRi-776, we compare our proposed DFR-ST with the following methods. 
Local Maximal Occurrence Representaion (LOMO) \cite{liao2015person} and Bag of Words with Color Name \cite{weijer2009learning} (BOW-CN) are based on hand-craft features which are first proposed for person re-ID \cite{liao2015person, zheng2015scalable}. 
FACT \cite{liu2016largescale} is based on the fusion with colors and attribute features. 
Hard-View-EALN \cite{lou2020embedding} and VAMI \cite{zhou2018viewpoint} impose an adversarial network between the generator and the discriminator to obtain more robust cross-view features. 
Moreover, SCCN-Ft + CLBL-8-Ft \cite{zhou2018vehicle} utilizes two networks to learn the local and global multi-view features.
OIN \cite{Wang2017Orientation} produces region masks based on the clustering of key points and establishes overall features using these region masks and global features. 
SPAN w/ CPDM \cite{chen2020orientationaware} detects different parts of vehicle images and then generates an attentive feature representation by aggregating the global and three-part attentive features together.
Besides, RNN-HA (ResNet) \cite{wei2018coarse} employs RNN to capture hierarchical dependencies for vehicle re-ID. 
HPGN \cite{shen2020exploring} proposes a pyramid of the spatial graph networks to handle multi-scale spatial features. 
PRN + ReRanking \cite{chen2019partition} explores partition strategies on three dimensions of feature maps to promote overall performance further. 
Moreover, GRF+GGL \cite{liu2020group} designs an efficient group-group loss to accelerate feature learning. QD-DLF \cite{zhu2020vehicle} defines pooling operations on four directions to compress basic features and concatenates them together as deep quadruple features. 
SAVER \cite{Khorramshahi2020devil} focuses on self-supervised attention mechanisms to improve vehicle re-ID algorithm. 
As for methods using spatio-temporal clues, ReID + query expansion \cite{lv2019vehicle} ensembles multiple informative features from several existing approaches and builds an acquisition module for vehicle locations and timestamps. 
FACT+Plate-SNN+STR \cite{liu2016deep} and PROVID \cite{liu2016provid} both use a license plate recognition module and a spatio-temporal module based on FACT \cite{liu2016largescale}.
OIN+ST \cite{Wang2017Orientation} constructs a spatio-temporal regularization module in addition to OIN \cite{Wang2017Orientation} while VAMI \cite{zhou2018viewpoint} + STR \cite{liu2016largescale} integrates a spatio-temporal similarity to the original model.
\begin{figure}[!t]
	\centering
	\includegraphics[width=2.7in]{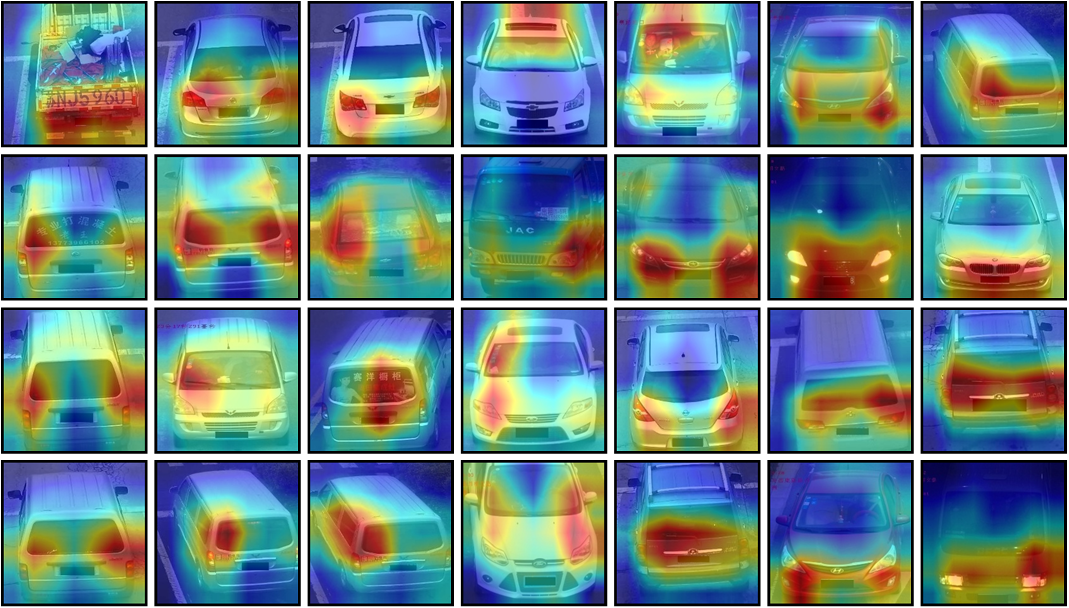}
			\setlength{\abovecaptionskip}{-3pt}
	\caption{Visualization of attention maps of DFR-ST on VehicleID. The red regions represent the attentive areas captured by the proposed network, which cover mostly car lights, car bumpers, sunroofs and car edges etc. 
	}
	\label{fig7-attention_visulization}
	\vspace{-15pt}
\end{figure}
Table \ref{table2:veri} illustrates the comparison results with state-of-the-art methods on VeRi-776. The methods listed in the upper part of Table \ref{table2:veri} only use visual appearance information while those appeared in the lower part of Table \ref{table2:veri} take both appearance and spatio-temporal cues into considerations. The proposed DFR-ST obtains the best Top-5 metric of 99.17\%  and achieves the highest mAP of 91.56\% and Top-1 of 97.74\% with the re-ranking strategy (DFR-ST + ReRanking). 

Firstly, compared with the approaches only based on visual appearance features, the proposed DFR-ST acquires the highest mAP of 91.56\% and the highest Top-1 of 97.74\%. This comparison verifies the intuition on the effectiveness of adding more information from other domains. Specifically, we notice that the Top-5 of PRN \cite{chen2019partition} is only a bit higher (0.23\%)  to our proposed DFR-ST. Although this observation indicates the advantage of the representation from \cite{chen2019partition}, we show that the unexpected noises frequent in unconstrained environments would degrade the performance of PRN \cite{chen2019partition} while our proposed DFR-ST maintain acceptable performance in the ablation studies. This comparison result could validate the superiority of the proposed DFR-ST. 

Secondly, among six vehicle re-ID methods based on multi-modal information (\emph{i.e.}, FACT + Plate-SNN + STR \cite{liu2016deep}, PROVID \cite{liu2016provid}, OIN + ST \cite{Wang2017Orientation}, Siamese-CNN + Path-LSTM \cite{shen2017learning}, VAMI \cite{zhou2018viewpoint} + STR \cite{liu2016largescale}, ReID + query expansion \cite{lv2019vehicle}), the proposed DFR-ST consistently outperforms FACT + Plate-SNN+STR \cite{liu2016deep}, PROVID \cite{liu2016provid}, OIN + ST \cite{Wang2017Orientation} and Siamese-CNN + Path-LSTM \cite{shen2017learning} by exceeding over 30\% on mAP, 12\% on Top-1 and 5\% on Top-5 at least. 
This comparison demonstrates that DFR-ST can take advantage of multi-modal information exceedingly from designed spatio-temporal scheme. 
Fig. \ref{fig4-veri-visualizer} visualizes the vehicle re-ID results on VeRi-776 dataset qualitatively. The bounding boxes in green describe the true positives, and those in red are false positives.

\subsection{Ablation Study}
Extensive experiments are conducted on two large-scale datasets, i.e., VeRi-776 and VehicleID, to thoroughly analyze each component's effectiveness of the proposed approach.

\begin{table}[!t]
	\renewcommand{\arraystretch}{1.3}
	\setlength{\abovecaptionskip}{-3pt}
	\caption{Component effectiveness analysis on VeRi-776.}
	\label{table5:component}
	\centering
	\begin{tabular}{cccc}
		\toprule
		Methods & mAP & Top-1 & Top-5\\
		\midrule
		Baseline & 65.72 & 78.10 &84.63  \\
		$\mathbf{x}_c$ & 70.53 & 84.55 & 88.76 \\
		$\mathbf{x}_c + \mathbf{x}_{\text{f1}}$ & 76.90 & 87.24 & 91.77 \\  
		$\mathbf{x}_c + \mathbf{x}_{\text{f1}} + \mathbf{x}_{\text{fdiv}}$ & 84.47 & 93.02 & 97.13 \\ 
		$\mathbf{x}_c + \mathbf{x}_{\text{f1}} + \mathbf{x}_{\text{fdiv}} $ + ST  &  $\mathbf{86.00}$  & $\mathbf{95.67}$ & $\mathbf{99.17}$ \\
		\bottomrule
	\end{tabular}
	\vspace{-10pt}
\end{table}

\subsubsection{Component Analysis}
We conduct ablation experiments on each component in the proposed DFR-ST to validate each component's effectiveness in the overall architecture. Table \ref{table5:component} shows the experimental results. 
The baseline adopts a ResNet-50 network with the cross-entropy loss. $\mathbf{x}_c$ denotes the features extracted from the coarse-grained feature stream. $\mathbf{x}_{\text{f1}}$ and $\mathbf{x}_{\text{fdiv}}$ represent the features from the first branch and the rest three-division branches of the fine-grained feature stream, respectively. Moreover, $ST$ is the spatio-temporal module. 
\begin{table}[!t]
	\renewcommand{\arraystretch}{1.3}
		\setlength{\abovecaptionskip}{-3pt}
	\caption{Effectiveness of spatio-temporal information on VeRi-776.}
	\label{table:stModuleEffect}
	\centering
	\begin{tabular}{cccc}
		\toprule
		Methods & mAP & Top-1 & Top-5\\
		\midrule
		FACT w/o STR \cite{liu2016largescale} & 19.92 & 59.65 & 75.27\\  
		+ STR \cite{liu2016largescale} & (+6.85\%) & (+1.79\%)& (+1.02\%) \\
		\hline
		OIN w/o ST \cite{Wang2017Orientation} & 48.00& 65.9 & 87.7 \\ 
		+ ST \cite{Wang2017Orientation}  &(+3.42\%) &(+2.40\%)& (+2.00\%) \\
		\hline
		VAMI \cite{zhou2018viewpoint} & 50.13 & 77.03 & 90.82 \\ 
		+ STR \cite{liu2016largescale} & (+11.19\%)&(+8.89\%) &(1.02\%) \\
		\hline
		Siamese-CNN \cite{shen2017learning} & 54.21& 79.32 & 88.92\\
		+ Path-LSTM \cite{shen2017learning}  &(+4.06\%)&(+4.17\%)&(+1.12\%)\\
		\hline
		DFR-ST w/o ST (ours) & 84.47 & 93.02 & 97.13 \\ 
		+ STR \cite{liu2016largescale} &(+1.72\%)&(+1.52\%)&(+0.67\%)\\
		+ ST (ours) &(+1.53\%) &(+2.65\%)& (+2.04\%)\\
		\bottomrule
	\end{tabular}
	\vspace{-15pt}
\end{table}

\begin{table*}[!t]
	\renewcommand{\arraystretch}{1.3}
	\setlength{\abovecaptionskip}{-3pt}
	\caption{Effectiveness of the appearance module on VehicleID.}
	\label{table3:vehicleid}
	\centering
	\begin{tabular}{@{}c|ccc|ccc|ccc@{}}
		\Xhline{1pt}
		\multirow{2}*{Methods} & \multicolumn{3}{c}{Test Size=800} & \multicolumn{3}{c}{Test Size=1600} &\multicolumn{3}{c}{Test Size=2400} \\
		\cline{2-10}
		& mAP & Top-1 & Top-5 & mAP & Top-1 & Top-5 & mAP & Top-1 &Top-5\\ 		
		\hline
		BOW-CN \cite{zheng2015scalable} & N/A &13.14&22.69&N/A& 12.94&21.09 &N/A& 10.20& 17.89\\
		LOMO \cite{liao2015person} & N/A & 19.74 &32.14 &N/A &18.95&29.46&N/A& 15.26& 25.63\\ 
		FACT + Plate-SNN \cite{liu2016deep} &49.2&43.62&64.84&44.8&39.94&62.98&38.6&35.68&56.24\\ 
		PROVID \cite{liu2016provid} & N/A& 48.90&69.51&N/A&43.64&65.34&N/A&38.63&60.72\\ 
		DRDL \cite{liu2016relative} & N/A & 49.0 & 73.5 & N/A & 42.8 & 66.8 & N/A & 38.2 & 61.6 \\
		DenseNet121 \cite{huang2017densely} & 68.85 & 66.10 &77.87 & 69.45 & 67.39 & 75.49 & 65.37 &63.07&72.57\\  
		TAMR \cite{guo2020twolevel} &N/A & 66.02& 79.71 &N/A&62.90&76.80&N/A&59.69&73.87\\ 
		VAMI \cite{zhou2018viewpoint} & N/A &63.12 &83.25 &N/A &52.87 &75.12 &N/A &47.34 &70.29 \\ 
		GS-TRE \cite{bai2018group} & 75.4 &75.9 &84.2 & 74.3 &74.8 &83.6 & 72.4 &74.0 &82.7 \\ 
		QD-DLF \cite{zhu2020vehicle} & $76.54$&72.32&92.48& $74.63$ &70.66 &88.90 &$68.41$ &64.14&83.37 \\ 
		RNN-HA (ResNet + 672) 
		\cite{wei2018coarse} & N/A &\underline{83.8} & 88.1 & N/A &$81.9$ & 87.0 & N/A &$ 81.1$ & 87.4 \\ 
		PRN (Single Height-Channel Branch) 
		\cite{chen2019partition}&N/A & 78.92 & 94.81&N/A & 74.94& $\underline{92.02}$ & N/A&71.58&88.46 \\ 
		GRF + GGL \cite{liu2020group} & N/A &77.1 &92.8 &N/A &72.7 &89.2 & N/A &70.0&87.1 \\ 
		Hard-View-EALN \cite{lou2020embedding} & 77.5 & 75.11& 88.09& 74.2& 71.78 &83.94 &71.0 &69.30 &81.42\\ 
		OIN \cite{Wang2017Orientation} &N/A&N/A&N/A&N/A&N/A&N/A&N/A&67.0&82.9\\ 
		SAVER \cite{Khorramshahi2020devil} & N/A&79.9 & $\underline{95.2}$ & N/A &77.6& 91.1& N/A& 75.3& 88.3\\ 
		HPGN \cite{shen2020exploring} &$\mathbf{89.60}$ &$\mathbf{83.91}$ & N/A& $\mathbf{86.16}$&$\mathbf{79.97}$&N/A&$\mathbf{83.60}$&\underline{77.32}&N/A\\ 
		\hline
		DFR-ST w/o ST (ours)   & 87.55&82.15 & $\mathbf{95.39}$ &\underline{84.94} &\underline{79.33}&$\mathbf{92.76}$& \underline{83.18}&$\mathbf{77.93}$ &$\mathbf{89.52} $ \\
		DFR-ST w/o ST (ours) + Re-Ranking & \underline{87.76} &82.71&95.02 &83.58& 77.66&91.28&82.28&76.96& \underline{88.48}\\
		\Xhline{1pt}
	\end{tabular}
	\vspace{-15pt}
\end{table*}

As depicted in Table \ref{table5:component}, the first observation is that only utilizing $\mathbf{x}_c$ for vehicle re-ID can improve the performance by 4.81\% on mAP comparing to the baseline, which confirms the useful combination of the cross-entropy loss and the triplet loss with the batch hard sampling strategy. 
After adding $\mathbf{x}_{\text{f1}}$, extracted from the first branch of the fine-grained feature stream, the mAP metric is enhanced by an additional 6.37\% and the Top-1, Top-5 metrics are increased by 2.69\%, 3.01\% respectively. This improvement indicates the aggregation of the coarse-grained features $\mathbf{x}_c$ and $\mathbf{x}_{\text{f1}}$ focusing more on local regions by AttentionNet, can provide extra useful information to recognize and identify the same vehicle. 
Furthermore, the participation of $\mathbf{x_{\text{fdiv}}}$ actively promotes the overall performance of the proposed method. Three metrics have improved by 7.57\%, 5.78\%, 5.36\%. This analysis confirms the effectiveness of division operations in different dimensions.
Moreover, $ST$ brings 1.53\% improvement on mAP and 2.65\%, 2.04\% on Top-1, Top-5, which verifies the complementary of spatio-temporal relationships with appearance features. 

\subsubsection{Effectiveness of the Appearance Module}

To validate the effectiveness of the proposed appearance module, we further compare our proposed DFR-ST with several current methods on VehicleID. 
Table \ref{table3:vehicleid} illustrates the mAP, Top-1, and Top-5 metrics of the proposed appearance module and state-of-the-art methods on the large-scale VehicleID dataset. The bold value means the first, and the underlined value stands for the second. 
Note that VehicleID dataset does not provide any other information besides images, hence we can not utilize the spatio-temporal module in the performance comparison.

As shown in Table \ref{table3:vehicleid}, the proposed appearance module outperforms all the existing methods on the Top-5 metric with small, medium, and large test sizes. Besides, the proposed DFR-ST also achieves the highest Top-1 of 77.93\% with the large test size, which is the most challenging setting among three test splits. 
In particular, compared with traditional hand-craft methods, LOMO \cite{liao2015person} and BOW-CN \cite{zheng2015scalable}, deep-learning-based approaches achieve significant improvements which verifies the strong representation capability of deep neural networks on the non-linear transformation. 


Furthermore, our proposed appearance module beats the existing methods including FACT \cite{liu2016largescale}, PROVID \cite{liu2016provid}, DRDL \cite{liu2016relative}, DenseNet121 \cite{huang2017densely}, TAMR \cite{guo2020twolevel}, VAMI \cite{zhou2018viewpoint},  GS-TRE \cite{bai2018group}, QD-DLF \cite{zhu2020vehicle}, PRN (Single Height-Channel Branch) \cite{chen2019partition}, GRF + GGL \cite{liu2020group}, Hard-View-EALN \cite{lou2020embedding}, OIN \cite{Wang2017Orientation} and SAVER \cite{Khorramshahi2020devil} on the mAP, Top-1 and Top-5 metrics with three test splits. 
Although RNN-HA (ResNet+672) \cite{wei2018coarse} receives more 1.65\% on Top-1 with the small test size than the proposed DFR-ST w/o ST, the proposed appearance module outperforms RNN-HA (ResNet+672) \cite{wei2018coarse} on the other metrics with three test sizes, which show the effectiveness of the proposed approach. 
Although our DFR-ST without the spatio-temporal module is a bit inferior to HPGN \cite{shen2020exploring}, the proposed DFR-ST can still obtain rather competitive performance on the challenging VehicleID dataset, which confirms  the competitiveness of our proposed appearance module. 

\subsubsection{Effectiveness of AttentionNet}
We conduct qualitative experiments to examine the effectiveness of multi-domain attention schemes in AttentionNet. Fig. \ref{fig7-attention_visulization} displays typical examples of the attention maps obtained by the proposed DFR-ST method for interpreting the multi-domain attention. Most of the activated regions (marked by the red color) are the car lights, car windows, the headstocks, and car edges. These regions can be interpreted by humans easily and have semantic meanings, which validates the fine-grained features' capability with the multi-domain attention mechanism to learn salient and informative regions for identifying near-duplicate vehicle images. 
This property benefits DFR-ST to maintain acceptable performance under image contamination, which is frequent and common in open and unconstrained environments.  

\begin{figure*}[!t]
	\centering
	\subfloat[]{\includegraphics[width=2.3in]{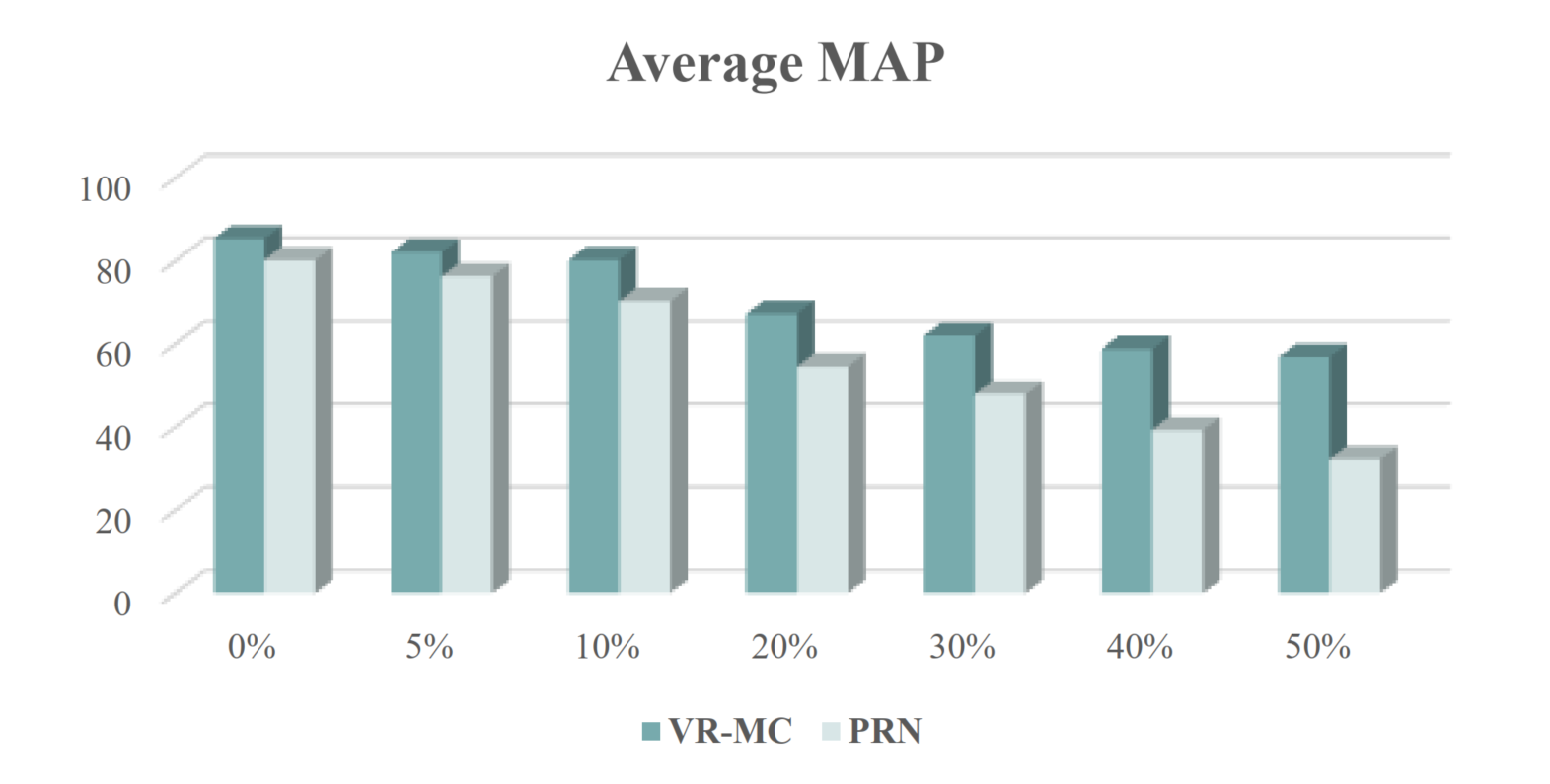}%
		\label{fig-prnMAP}}
	\hfil
	\subfloat[]{\includegraphics[width=2.3in]{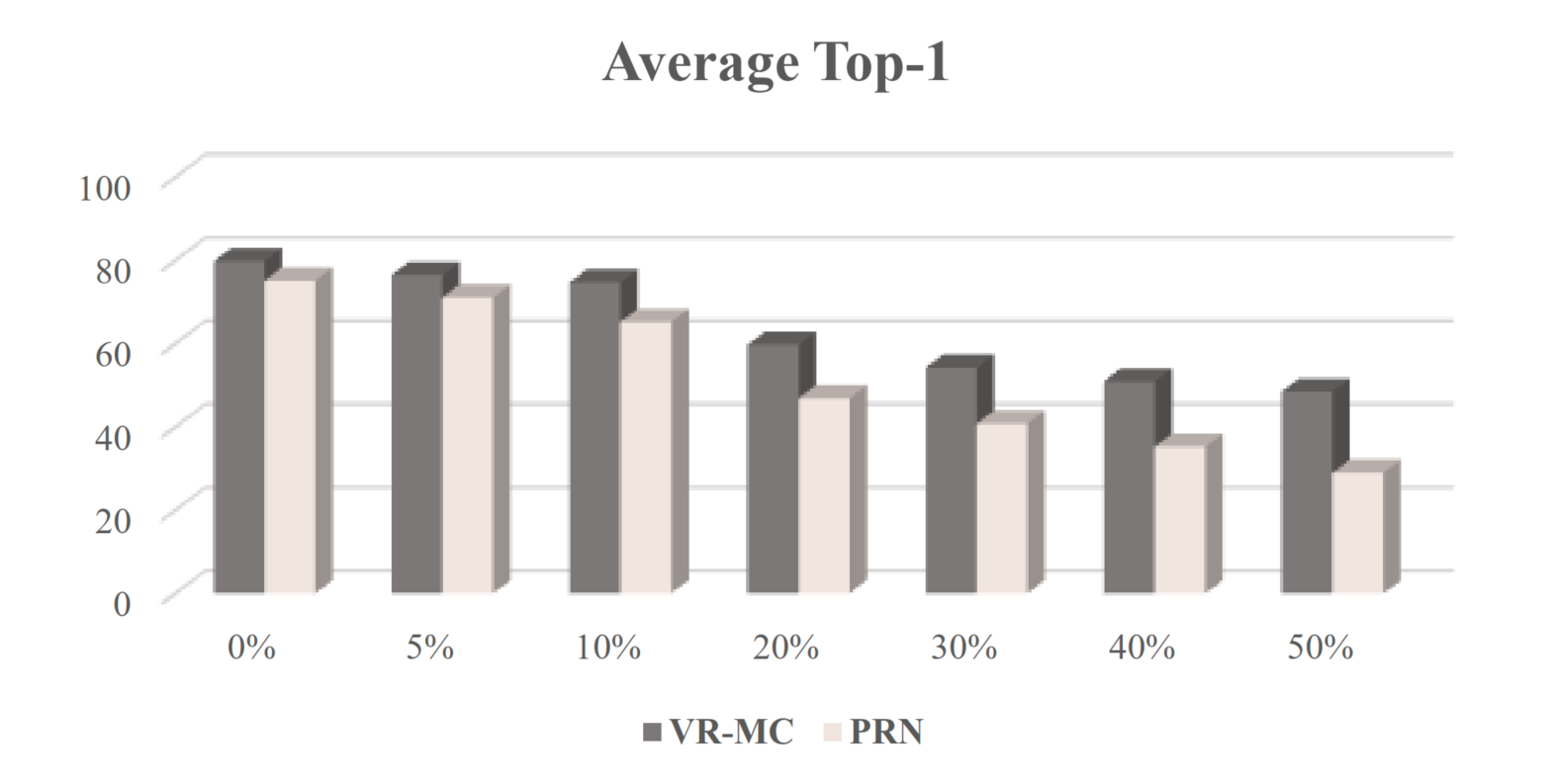}%
		\label{fig-prnTOP1}}
	\hfil
	\subfloat[]{\includegraphics[width=2.3in]{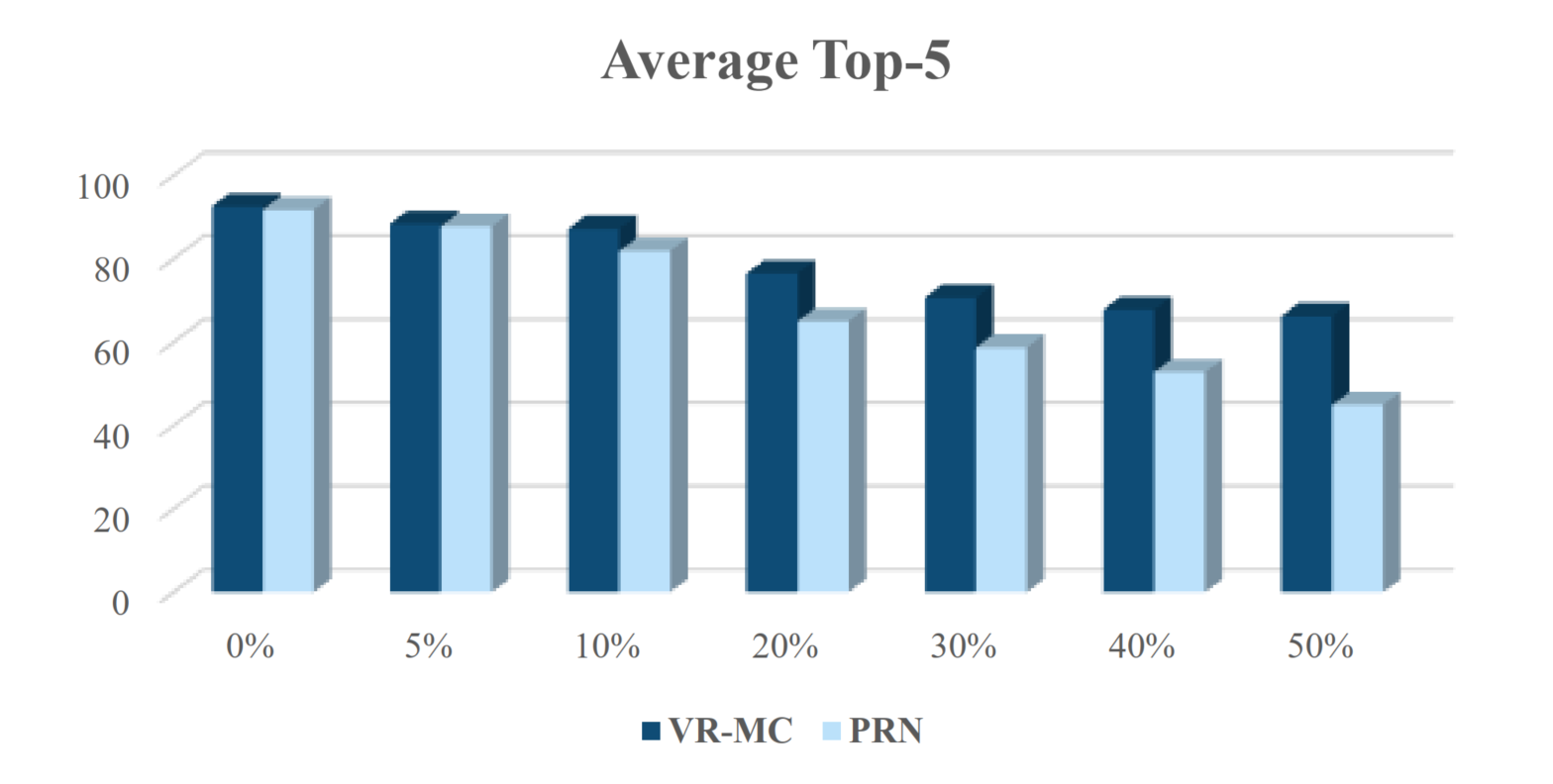}%
		\label{fig-prnTOP5}}
	\caption{Performance comparision of the proposed DFR-ST and PRN \cite{chen2019partition} with different percentages of image noises. 
		With the percentage of the image contamination becomes larger, the proposed DFR-ST obtains a slower rate of performance degradation than PRN \cite{chen2019partition}. 
	}
	\label{fig-prnNoise}
	\vspace{-10pt}
\end{figure*}

\begin{table*}[!t]
\renewcommand{\arraystretch}{1.3}
	\setlength{\abovecaptionskip}{-3pt}
\caption{Performance evaluation of DFR-ST (ours) with different contamination percentages on VehicleID.}
\label{table4:contamination}
\centering
\begin{tabular}{@{}c|ccc|ccc|ccc|ccc@{}}
\Xhline{1pt}
\multirow{2}*{Contaminated Percentages} & \multicolumn{3}{c}{Test Size=800} & \multicolumn{3}{c}{Test Size=1600} &\multicolumn{3}{c}{Test Size=2400} & \multicolumn{3}{c}{Average}\\
\cline{2-13}
& mAP & Top-1 & Top-5 & mAP & Top-1 & Top-5 & mAP & Top-1 &Top-5& mAP & Top-1 &Top-5\\ 		
\hline
0\%   & 87.76 &82.15& 95.39&84.94 &79.33&92.76& 83.18&77.93&89.52 &85.29&79.80&92.56\\
5\% &84.47 &78.83&91.41 &81.50&75.99&88.32&80.09&74.77&85.48 &82.02&76.53&88.40\\
10\% &82.55&76.92&89.71&79.36&73.72&86.88&78.72&73.49&85.04&80.21&74.71&87.21\\
20\% &69.18&61.13&78.73& 68.35&61.37&76.98&64.38&56.63&73.70 &67.30&59.71&76.47\\
30\% &65.75&58.62&73.69&61.53&53.59&71.10&58.24&50.46&67.29&61.84&54.22&70.69\\
40\% &61.61&54.12&70.35&59.77&51.84&69.21&54.81&46.84&63.73&58.73&50.93&67.76\\
50\% &60.66 &52.47 &69.93 &56.06 &47.75 &65.64 &53.70 &45.35 & 63.24&56.81&48.52&66.27\\
\Xhline{1pt}
\end{tabular}
\vspace{-12pt}
\end{table*}

\begin{table*}[!t]
\renewcommand{\arraystretch}{1.3}
	\setlength{\abovecaptionskip}{-3pt}
\caption{Performance evaluation of PRN \cite{chen2019partition} with different contamination percentages on VehicleID.}
\label{table:prn}
\centering
\begin{tabular}{@{}c|ccc|ccc|ccc|ccc@{}}
\Xhline{1pt}
\multirow{2}*{Contaminated Percentages} & \multicolumn{3}{c}{Test Size=800} & \multicolumn{3}{c}{Test Size=1600} &\multicolumn{3}{c}{Test Size=2400} & \multicolumn{3}{c}{Average}\\
\cline{2-13}
& mAP & Top-1 & Top-5 & mAP & Top-1 & Top-5 & mAP & Top-1 &Top-5& mAP & Top-1 &Top-5\\ 		
\hline
0\%	& 84.08 & 78.92 & 94.81& 80.54 & 74.94& 92.02 & 75.76 &71.58&88.46 & 80.13  &  75.15& 91.76\\ 
5\% & 80.53 & 74.38 & 91.28 &76.90 & 70.86 & 88.46 & 71.49& 68.10 &83.61&76.31&71.11&87.78\\
10\% & 73.37 & 69.81 & 84.24& 71.20 &64.50 &81.88& 66.45& 61.32& 79.44&70.34&65.21&81.85\\
20\% &58.87 & 50.60& 67.12& 54.40& 46.73& 65.03& 50.38 & 43.18& 62.89&54.55&46.84&65.01\\
30\% &50.47 & 43.91 & 61.54 &47.62 &40.86&58.17& 45.59&37.63&56.40&47.89&40.80&58.70\\
40\% &43.71&38.30&55.99& 38.53 &35.66 & 52.25& 35.48&31.70&50.06&39.24&35.22&52.77\\
50\% &35.69& 32.42& 48.16&31.80&28.75&44.26&29.48&25.74&42.39&32.32&28.97&44.94\\
\Xhline{1pt}
\end{tabular}
\vspace{-10pt}
\end{table*}

\subsubsection{Performance Evaluation with Image Contamination}
In real-world scenarios, images are inevitably contaminated with noises due to unreliable sensing devices, limited network communication resources, or time-variant transmission environments. Thus, we conduct additional experiments to test the performance of the proposed DFR-ST under unexpected noises to demonstrate its robustness. 
We simulate three main problems (\emph{i.e.}, mosaics, color cast, and abnormal brightness) caused by unreliable transmission as noise categories. 
The percentage of noisy pixels measures the degree of image contamination to total pixels. Note that YUV pixel matrices' improper decoding produces these noises instead of the decoding of elementary streams with MPEG-2 proposal.

As shown in Table \ref{table4:contamination}, firstly, the proposed DFR-ST performance only drops 5.08\% on mAP, 5.09\% on Top-1, and 5.35\% on Top-5 with 10\% noisy pixels on average. This result confirms the robustness of the proposed DFR-ST. Note that DFR-ST with even 10\% image contamination still obtains competitive performance compared to QD-DLF \cite{zhu2020vehicle} and defeats GS-TRE \cite{bai2018group}, DJDL \cite{liu2019supervised}, VAMI \cite{zhou2018viewpoint} and DenseNet121 \cite{huang2017densely}. This comparison validates the effectiveness of DFR-ST.

Secondly, we notice that the performance of DFR-ST gets worse with larger image noises. However, even with the worst performance, i.e., under 50\% noisy pixels, DFR-ST still has a better performance on VehicleID than several methods including DRDL \cite{liu2016relative}, FACT \cite{liu2016largescale}, LOMO \cite{liao2015person} and BOW-CN \cite{zheng2015scalable}. This result verifies DFR-ST can maintain an acceptable performance with a severe degree of image contamination, which is advantageous for real-world applications.


Besides, we reproduce PRN \cite{chen2019partition} and compare its capability to resist image contamination with the proposed DFR-ST on VehicleID. For the fairness of the comparison, PRN (Single height-channel branch) is evaluated thoroughly with DFR-ST w/o ST module because PRN (Single height-channel branch) has better performance than the complete PRN. The comparison results are shown in Fig. \ref{fig-prnNoise} and Table \ref{table:prn}. With the image noises grow larger, the performance of PRN \cite{chen2019partition} decreases faster than the proposed DFR-ST. This comparison validates the robustness of the proposed DFR-ST to resist image noises.

\section{Conclusion}
\label{section:conclusion}

In this paper, we proposed a discriminative feature representation DFR-ST with multi-modal cues for vehicle re-ID. The scheme consists of an appearance module for establishing robust visual features integrating coarse-grained and fine-grained features, and a spatio-temporal module for constructing a quantitative description of the transition distances of camera locations and time snapshots. Extensive experiments on two large-scale datasets validate the superiority of the proposed DFR-ST and the complementary of appearance and spatio-temporal information. 
In the future, we will exploit the improvement of more discriminative representation for vehicle re-ID and real-time video analysis. 

\ifCLASSOPTIONcaptionsoff
  \newpage
\fi



\bibliographystyle{IEEEtran}
\bibliography{mybibfile}

\appendices
\section{Effect of the Division Branches}
Table \ref{table7:slicing} presents the evaluation results of variants on the division branches on the VeRi-776 dataset. The first variant refers to the appearance module without three division branches, which only contains a dual path of the coarse-grained features stream and the first branch of the fine-grained feature stream. 

The first observation of this comparison is that the division operations along three dimensions contribute to the overall performance, and the channel division branch brings the most significant improvement among these three categories of division branches. This result is also observed in work PRN \cite{chen2019partition}, which only uses a single channel branch can get the highest performance on VehicleID dataset than the fusion of three branches from different dimensions. 
Different from PRN \cite{chen2019partition}, we discover that the combination of multiple primary division branches can promote the performance further. Besides, we achieve the best performance by employing these three division strategies on VeRi-776. It indicates the complementary of the information provided from the division strategies along different axes. 

\section{Effect of placements in AttentionNet}
We conduct extensive experiments to investigate the effect on different placements of the multi-domain attention scheme in the fine-grained feature stream of the proposed DFR-ST on VeRi-776 dataset. Table \ref{table6:attention-order} shows the performance with different orders of the channel attention and spatial attention. Note that the experiments neglect the spatio-temporal module for simplicity. 
As observed in Table \ref{table6:attention-order}, the consolidation of the awareness in different domains (i.e., the spatial domain and the channel domain) can promote the overall performance rather than single-domain attention from the experimental results. 
Moreover, we adopt the placement of the sequential channel attention and spatial attention for better performance in further experiments. However, the empirical consequences reveal that the order of the channel attention and spatial attention is insubstantial to the overall performance. 

\begin{table}[!t]
	\renewcommand{\arraystretch}{1.3}
	\caption{Performance analysis of the division branches on VeRi-776.}
	\label{table7:slicing}
	\centering
	\begin{tabular}{cccc}
		\toprule
		Variants & mAP & Top-1 & Top-5\\
		\midrule
		w/o division & 76.90 & 87.24 & 91.77  \\
		Height division & 77.81   & 85.29  & 89.59 \\
		Width division &  78.28  & 87.43  &  91.31  \\
		Channel division & 82.92   & 90.46  &  94.06 \\
		Height + Width division & 82.67   & 92.14 & 94.89   \\
		Height + Channel division & 84.30    & 92.52  & 96.10   \\
		Width + Channel division & 83.82   & 92.35   & 96.96   \\
		Height + Width + Channel division & $\mathbf{84.47}$ & $\mathbf{93.02}$ & $\mathbf{97.13}$   \\
		\bottomrule
	\end{tabular}
	\vspace{-10pt}
\end{table}
\begin{table}[!t]
	\renewcommand{\arraystretch}{1.3}
	\caption{The Effect of different placements of attention sub-modules in AttentionNet on VeRi-776.}
	\label{table6:attention-order}
	\centering
	\begin{tabular}{cccc}
		\toprule
		Placements & mAP & Top-1 & Top-5\\
		\midrule
		Channel Attention & 83.69 & 90.17 & 95.04 \\
		Spatial Attention & 82.55 & 90.76 & 94.48 \\
		Sequential Channel + Spatial Attention  & $\mathbf{84.47}$ & $\mathbf{93.02}$ & $\mathbf{97.13}$ \\
		Sequential Spatial + Channel Attention  & 83.74 &  92.81  & 96.35 \\
		Parallel Spatial + Channel Attention  &  83.51 &  92.38  & 95.86\\
		\bottomrule
	\end{tabular}
\end{table}

\section{Parameter Setting}
We first investigate the influence of $\lambda$ in Eq. (\ref{eq:loss}). As shown in Table \ref{table:lambda}, we set $\lambda = 0.4$ in all experiments because adding the triplet loss can boost the overall performance further. 

\begin{table}[!t]
	\renewcommand{\arraystretch}{1.3}
	\caption{Evaluation on the influence of $\lambda$ on VeRi-776.}
	\label{table:lambda}
	\centering
	\begin{tabular}{cccc}
		\toprule
		$\lambda$ & mAP & Top-1 & Top-5\\
		\midrule
		0   & 83.04   & 91.55   & 96.18  \\
		0.1 & 83.56 & 92.32 & 96.70  \\
		0.2 & 84.08  & 92.64  & 96.82 \\ 
		0.3 & 84.41 & $\mathbf{93.05}$ & 97.00   \\
		0.4 & $\mathbf{84.47}$ & 93.02 & $\mathbf{97.13}$   \\
		0.5 & 84.43 & 92.94 & 96.89   \\
		0.6 & 84.30 & 92.96  &  96.80 \\
		0.7 & 84.15 & 92.74  & 96.53 \\  
		0.8 & 84.07 & 92.43  & 96.45 \\
		0.9 & 83.69 & 92.16  & 96.39 \\
		1.0 &  83.35  & 91.86  & 96.24 \\            
		\bottomrule
	\end{tabular}
\end{table}

\begin{table}[!t]
	\renewcommand{\arraystretch}{1.3}
	\caption{Evaluation on the influence of $\omega$ on VeRi-776.}
	\label{table:omega}
	\centering
	\begin{tabular}{cccc}
		\toprule
		$\omega$ & mAP & Top-1 & Top-5\\
		\midrule
		0   & 84.47 & 93.02 & 97.13 \\
		0.1 & 85.73 & 94.60 & 98.52  \\
		0.2 & $\mathbf{86.00}$ & $\mathbf{95.67}$ & 99.17 \\
		0.3 & 85.86 & 95.60  & $\mathbf{99.23}$ \\
		0.4 & 85.56 & 95.12  & 98.35 \\
		0.5 & 83.40 & 93.66  & 96.67  \\
		0.6 & 81.58   & 92.31  & 94.83 \\
		0.7 & 80.09   & 91.24  & 93.37 \\  
		0.8 & 78.35   & 88.28  & 92.54 \\
		0.9 & 75.70   & 84.35  & 89.11 \\
		1.0 & 73.61   & 82.33  & 86.64 \\            
		\bottomrule
	\end{tabular}
	\vspace{-10pt}
\end{table}

\begin{figure}[!t]
	\centering
	\subfloat[]{\includegraphics[width=1.5in]{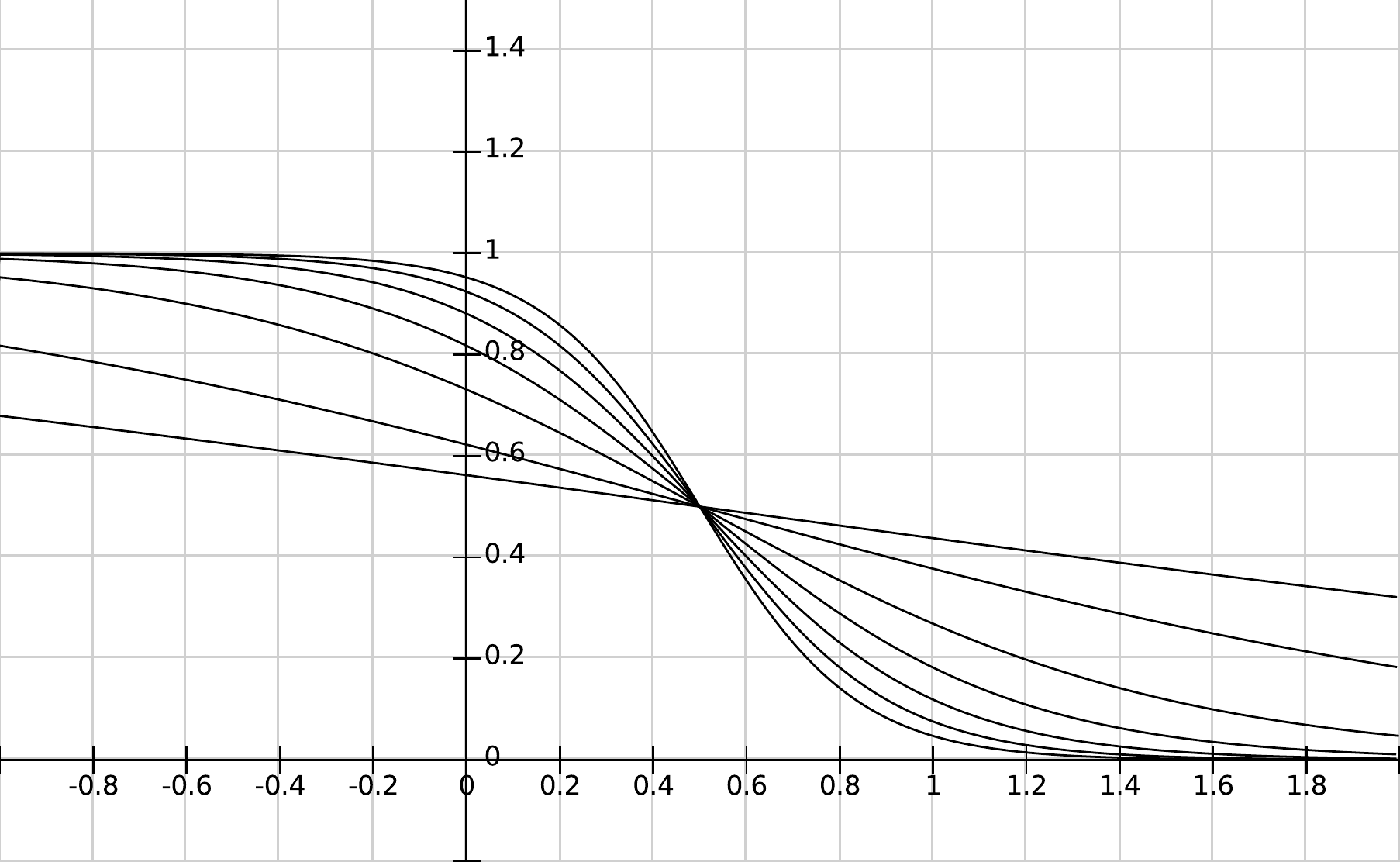}%
		\label{fig9-alpha}}
	\hfil
	\subfloat[]{\includegraphics[width=1.5in]{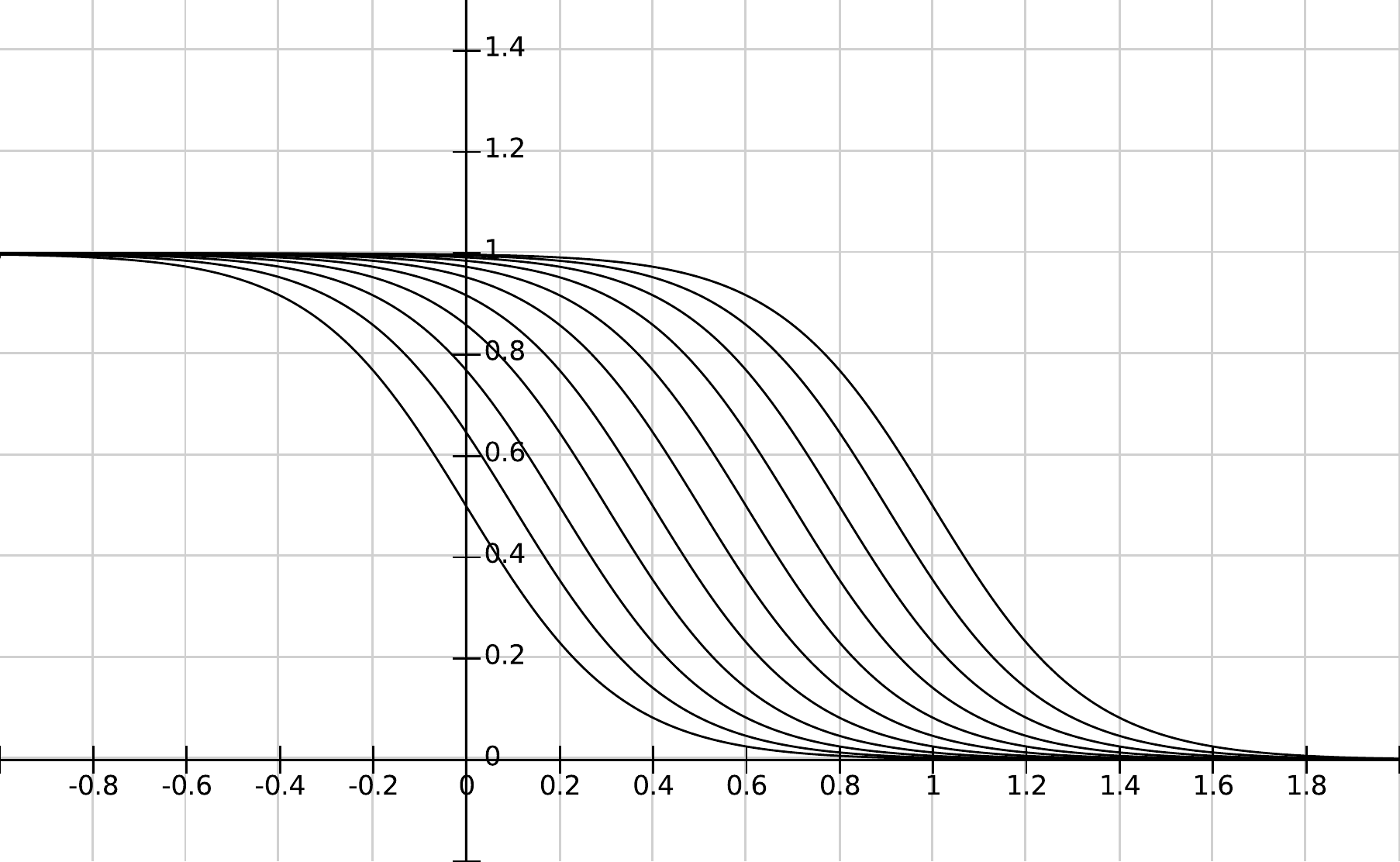}%
		\label{fig10-beta}}
	\caption{Visualization the effect of $\alpha$ and $\beta$ on the distribution shapes. 
		(a) The effect on the distribution shapes of the parameter $\alpha$. The curves are $D = \frac{1}{1+\exp (\alpha (x-0.5))}$ as $\alpha =0.5, 1, 2, \cdots, 6$ based on the ascending order of the intersections across the y axis.  
		(b) The effect on the distribution shapes of the parameter $\beta$. The curves are $D = \frac{1}{1+\exp (6 (x-\beta))}$ as $\beta =0, 0.1, 0.2, \cdots, 1$ from left to right.}
	\label{fig11-alpha-beta}
\end{figure}

Besides, we study the impact of $\omega$ in Eq. (\ref{eq:distance}), i.e., the weight of the spatio-temporal similarity between the image pairs in the distance measurement of Algorithm \ref{alg}.  Table \ref{table:omega} shows the experimental results. 
We first observe that a moderate $\omega$ can increase the performance, which validates the complementary of the spatio-temporal information and the appearance representation. 
Second, the performance becomes worse quickly as $\omega$ becomes larger. It shows that too large weights on spatio-temporal cues can bring unexpected degradation, and the appearance features account for more contributions to the proposed vehicle re-ID approach. This result has following the human perception of the world, in which the vision accounts for over 80\% information. Therefore, we set $\omega = 0.2$ for all experiments.

Moreover, we investigate the effect of different distribution shapes controlled by $\alpha_1$, $\alpha_2$ and $\beta_1$, $\beta_2$ in the spatio-temporal module. 
The distribution in Eq. \ref{eq:alpha} delineates the declining speeds of the spatio-temporal similarity with the differences of the camera locations and the timestamps between two vehicle images. The intuition is that we hope to maintain more plausible samples of high spatio-temporal similarities for the appearance module to make decisions and get rid of those relatively dissimilar images to reduce the search complexity. Thus, we set the the parameters $\alpha_1$ and $\alpha_2$ to be 6 and $\beta_1$, $\beta_2$ to be 0.5 based on Fig. \ref{fig11-alpha-beta}.





%

\end{document}